\documentclass{article}

\usepackage[preprint]{neurips_2025}

\usepackage[utf8]{inputenc} 
\usepackage[T1]{fontenc}    
\usepackage{hyperref}       
\usepackage{url}            
\usepackage{booktabs}       
\usepackage{amsfonts}       
\usepackage{nicefrac}       
\usepackage{microtype}      
\usepackage{xcolor}         

\usepackage{graphicx}
\usepackage{pifont}
\usepackage{multirow}
\usepackage{amsmath}
\usepackage{wrapfig}
\usepackage{caption}  
\usepackage{lipsum} 
\usepackage{floatflt}
\usepackage{subfigure} 
\usepackage{tcolorbox}
\usepackage{bm} 
\tcbuselibrary{breakable}
\usepackage{algorithm}      
\usepackage{algorithmic}

\title{FAS: Fast ANN-SNN Conversion for Spiking Large Language Models}

\author{%
  Long Chen$^1$, Xiaotian Song$^1$, Andy Song$^2$, BaDong Chen$^3$, Jiancheng Lv$^1$, Yanan Sun$^1$ \\
  $^1$Collage of Computer Science, Sichuan University\\
  $^2$School of Computing Technologies, Royal Melbourne Institute of Technology University\\
  $^3$Institute of Artificial Intelligence and Robotics, Xi’an Jiaotong University\\
}

\begin{document}

\maketitle
 
\begin{abstract}
Spiking Large Language Models have been shown as a good alternative to LLMs in various scenarios.   Existing methods for creating Spiking LLMs, i.e., direct training and ANN-SNN conversion, often suffer from performance degradation and relatively high computational costs.  To address these issues, we propose a novel Fast ANN-SNN conversion strategy (FAS) that transforms LLMs into spiking LLMs in two stages. The first stage employs a full-parameter fine-tuning of pre-trained models, so it does not need any direct training from scratch.  The second stage introduces a coarse-to-fine calibration method to reduce conversion errors and improve accuracy. Experiments on both language and vision-language tasks across four different scales of LLMs demonstrate that FAS can achieve state-of-the-art performance yet with significantly reduced inference latency and computational costs. Notably, FAS only takes eight timesteps to achieve an accuracy of 3\% higher than that of the OPT-7B model, while reducing energy consumption by 96.63\%.\footnote{Available code:~https://github.com/lc783/FAS}.
\end{abstract}

\section{Introduction}
Large Language Models (LLMs), with recent success in various applications, e.g., GPT-3~\cite{brown2020language}, LLaVA~\cite{liu2024visual} and LLaMA 3~\cite{dubey2024llama}, have become strong candidates in various tasks. However, all these models suffer from high energy consumption, mainly due to Floating-Point Multiplication and Addition (MAC) operations. For example, training GPT-3 consumes $\sim$1,287 MWh of energy~\cite{deVries2023TheGE}, which is equivalent to the annual energy consumption of 120 households.
In recent years, a low-power alternative to vanilla LLMs, Spiking LLMs, have appeared, which are based on Spiking Neural Networks (SNNs), inspired by the spiking signalling mechanism of brain neurons~\cite{bal2024spikingbert,zhu2023spikegpt}. Through the energy-saving strategies of SNNs, i.e., computing with discrete binary spikes, the floating-point MAC of LLMs can be significantly reduced in Spiking LLMs, hence achieving low cost~\cite{davies2018loihi,yao2024spikeChip}.

Existing SNN-based approaches can be categorized into direct training and ANN-SNN conversion. The former typically uses backpropagation with surrogate gradient~\cite{8891809, zenke2021remarkable,lian2023learnable}. These approaches require training an SNN model from scratch, which is inherently time-consuming and resource-consuming, especially for LLMs.  Thus, the most common practice of these approaches only focus on training some components of LLMs on simple  tasks~\cite{yao2024spike,song2024one}.  On the contrary, ANN-SNN conversion~\cite{cao2015spiking, rueckauer2016theory,rueckauer2017conversion} aims to convert ANN's analog neurons to spiking neurons, while eliminating the errors caused by the conversion.  In this way, good-performing SNNs can be obtained with less or without a training process~\cite{Deng2021OptimalCO,Li2022EfficientAA}. However, existing ANN-SNN conversion methods cannot handle LLMs effectively, because the training strategy and scale of LLMs are significantly different from those of Recurrent Neural Networks (RNNs) and Convolutional Neural Networks (CNNs).


\begin{wrapfigure}{r}{0.45\columnwidth}
  \centering
    \includegraphics[width=\linewidth]{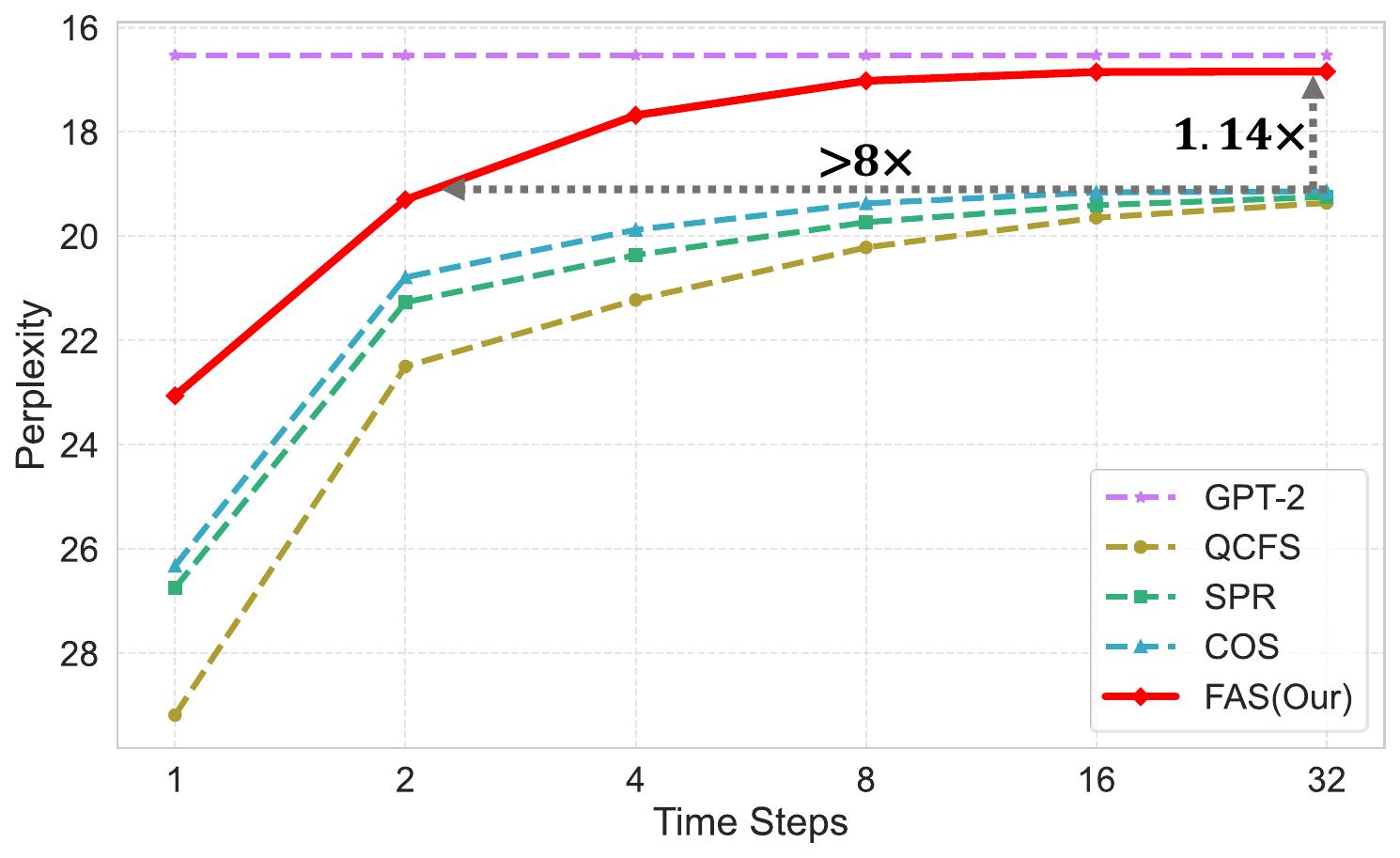}
    \caption{Performance of ANN-SNN conversion methods on GPT-2 for WikiText-103.}
  \label{fig:compare}
\end{wrapfigure}
When dealing with LLMs, several problems arise.  
First, the training cost of LLMs is significantly higher than that of CNNs, making existing training-from-scratch conversion methods ineffective in the scenario. This is because existing NN-SNN conversion methods do not fully utilize the pre-trained weights~\cite{hao2023reducing, wang2023toward}. As a result, they lead to high computational intensity when applied to LLMs.
Second, the temporal errors in conversion on LLMs are larger than that on CNNs~\cite{bu2023optimal, hao2023reducing,Hao2023BridgingTG}, thereby resulting in severe performance degradation and inference latency. The reason is that the temporal errors accumulate layer by layer, and the deeper and larger LLMs suffer more temporal errors. As illustrated in Figure~\ref{fig:compare}, SOTA conversion methods suffer from a large performance gap, especially under few timesteps.


In this work, we propose a Fast two-stage ANN-SNN Conversion method, termed FAS, for spiking LLMs. Specifically, the first stage involves a fine-tuning of the pre-trained LLMs. This can eliminate the time-consuming process of training from scratch.  In the second stage, a coarse-to-fine calibration method is introduced to eliminate temporal errors.  Our contributions are summarized as follows:
\begin{itemize}
    \item We propose FAS, a two-stage ANN-SNN conversion method for spiking LLMs. It achieves high-performance conversion with a low computational cost.
    \item We reveal that temporal errors are severe in conversion for spiking LLMs. To address this, a novel coarse-to-fine calibration component is introduced.
    \item We conduct experiments on both language and vision-language tasks across three LLMs and one multimodal LLM, showing that FAS can achieve SOTA performance at low cost.
\end{itemize}

\section{Related Works}

\subsection{One-stage ANN-SNN Conversion} 
One-stage ANN-SNN conversion aims to directly convert ANN models to SNN models without any further optimization on the converted SNN models, e.g., fine-tuning. This type of method focuses on reducing the conversion as much as possible. For instance, Cao et al.~\cite{cao2015spiking} initially introduced the one-stage method by training ANNs with ReLU activation functions and then replacing these activations with spiking neurons. Based on this, Diehl et al.~\cite{Diehl2015FastclassifyingHS} proposed model-based and data-based normalization to narrow the gap between ANN and SNN. Furthermore, Sengupta et al.~\cite{Sengupta2018GoingDI} introduced scaling methods to normalize weights and thresholds of SNNs, improving the conversion performance. To further mitigate conversion loss, Rueckauer et al.~\cite{rueckauer2016theory} and Han et al.~\cite{han2020deep} introduced a ``reset-by-subtraction'' mechanism, which can enhance precision during conversion. More recently, Bu et al.~\cite{Bu2022OptimizedPI} analyzed conversion error and proposed the QCFS activation function to replace ReLU in ANNs, which can effectively reduce conversion loss. One-stage ANN-SNN conversion has shown promising performance, however, it typically requires a large number of time steps to achieve SOTA performance. Therefore, it is impractical to apply the one-stage methods to large models like LLMs.




\subsection{Two-stage ANN-SNN Conversion}
Two-stage ANN-SNN conversion involves additionally optimizing the SNN converted by the one-stage methods to further to improve its performance. For example, SPR~\cite{hao2023reducing} proposed an optimization strategy that uses residual membrane potential to reduce unevenness errors for converted SNN models. Similarly, COS~\cite{Hao2023BridgingTG} optimized the converted SNN models by shifting the initial membrane potential. However, SPR and COS require additional time steps to gather necessary prior information, which can reduce efficiency. To address this, LTL~\cite{yang2022training} introduced a local tandem learning rule, which can efficiently guide the training of the converted SNN models. In addition, EAC~\cite{li2024error} proposed a layer-wise calibration algorithm to optimize the converted SNN models. However, LTL and EAC need to optimize all parameters of the converted SNN model. In comparison, the proposed FAS method only optimizes the membrane threshold and initial membrane potential. FAS is simpler and more effective than other peer competitors, especially for LLMs.

\section{Preliminary}
\subsection{Analog Neuron Model for LLMs}

\noindent LLMs are structured layer by layer.  The output \(a^l\) of the neurons in the \(l\)-th layer is achieved through a linear weighted combination followed by a nonlinear mapping:
\begin{equation}
\label{eq1}
    a^l=f(W^la^{l-1}),
\end{equation}
where \(W^l\) is the weight matrix of the \(l\)-th layer, and \(f(\cdot)\) is the nonlinear activation function.

\subsection{Spiking Neuron Model}
\label{sec:spiking_neuron}
For SNN, we follow the conversions~\cite{Diehl2015FastclassifyingHS,Han2020RMPSNNRM,Deng2021OptimalCO} and consider the Integrate-and-Fire (IF) neuron model~\cite{cao2015spiking}. Its kinetic behavior can be represented by Eq.~(\ref{eq:vt}):
\begin{equation}
\label{eq:vt}
   v^{l}(t) =  v^{l}(t-1)+W^{l}S^{l-1}(t)\theta ^{l-1} -S^{l}(t)\theta ^{l},
\end{equation}
where \(v^{l}(t)\) represent the membrane potential  at time steps \(t\) in the \(i\)-th layer. \(W^{l}\) and \(\theta^l\) are the weight matrix and firing threshold of the IF neuron, respectively. \(S^{l}(t)\) denotes discrete spikes at the \(l\)-th layer at time steps \(t\). 
Note that when \(v^{l}(t-1)+W^{l}S^{l-1}(t)\theta ^{l-1}\) exceeds the threshold \(\theta^{l}\), the IF neuron is activated, so \(S^{l}(t)\) equals $1$. Otherwise, the IF neuron is inhibited and \(S^{l}(t)\) equals $0$.

\subsection{Conversion Error of ANN-to-SNN}
\label{sec:conver_error}
ANN-SNN conversion aims to establish a consistent relationship between the analog neurons and the spike rates of IF neurons. The spike rate $r^{l}(T)$ can be represented as Eq.~(\ref{eq9}) (see \textbf{Appendix}~\ref{sec:spike_rate}):
\begin{equation}
\label{eq9}
    r^{l}(T)=\text{clip}(\frac{\theta^{l}}{T}\left \lfloor \frac{TW^{l}r^{l-1}(T)+v^{l}(0)}{\theta ^{l}}  \right \rfloor, 0, \theta ^{l} ).
\end{equation}

Specifically, the conversion can be achieved by mapping the activation value \(a^l\) of ANNs (see Eq.~(\ref{eq1})) to the spike rate \(r^l\) of SNNs (see Eq.~(\ref{eq9})). However, the conversion process still has three types of conversion errors: \textit{\textcircled{\scriptsize{1}} Quantization error:} The spike rate is a discrete distribution, with values occurring at regular intervals of \(\theta/T\).  When \(a^{l} \in \left[k\theta^l/T, (k+1)\theta^l/T\right]\), it is mapped to \(k\theta^l/T\). The discrepancy \(a^l-k\theta^l/T\) is a source of errors. \textit{\textcircled{\scriptsize{2}} Clipping error:} This is caused by the different value ranges of ANN and SNN. Specifically, when \(a^l \in [0, a_{max}]\) and \(r^{l} \in [0, \theta^l]\), where \(a_{max}\) denote the max value in \(a^l\), the value  \(a^l \in [\theta^l, a_{max}]\) will be all mapped to \(\theta^l\), also generating errors.

\textit{\textcircled{\scriptsize{3}} Temporal error:} It refers to the inconsistency between \(a^l\) and \(r^l\) due to the fluctuation in the 
temporal sequences of spike arrivals in activation layers. This variation can result in a higher or lower number of spikes than expected, resulting in 
poor performance.
\begin{wrapfigure}{r}{0.5\columnwidth}
  \centering
  \vspace{15pt}
    \includegraphics[width=\linewidth]{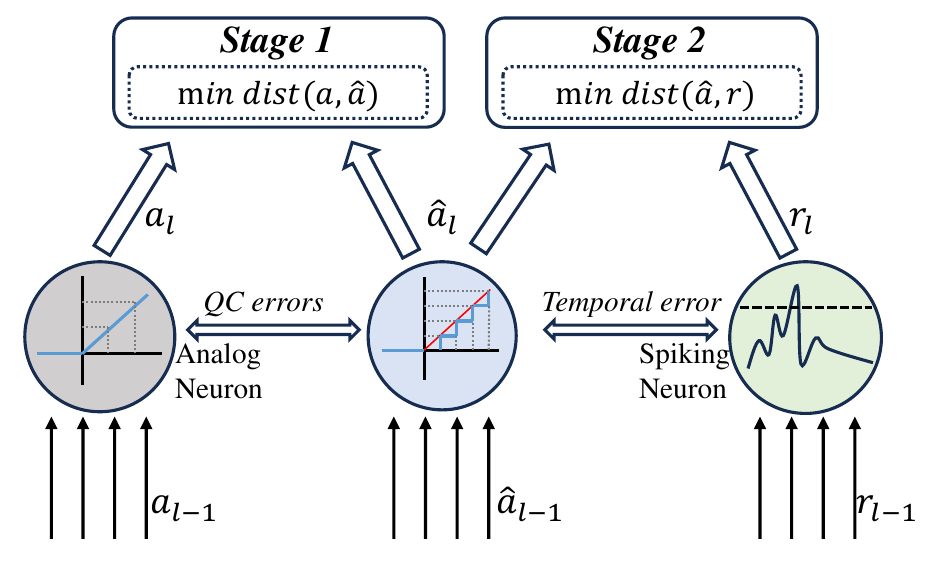}
    \caption{The overall framework of the proposed FAS method. \textit{QC errors} is composed of the \textit{quantization error} and the \textit{clipping error}.}
  \label{fig3}
  \vspace{-50pt}
\end{wrapfigure}
\section{Methodology}
\subsection{Overall Framework}

As discussed in Section~\ref{sec:conver_error}, the \textit{quantization errors} and \textit{clipping errors} come from the process of discretizing the continuous ANN activation function, and the \textit{temporal errors} are caused by using the disordered temporal sequences to generate spikes. The proposed FAS method is tailored for eliminating these errors in LLMs, which can secure the high-performance and low-energy spiking LLMs. Specifically, the overall framework of FAS is shown in Figure~\ref{fig3}, consisting of two stages. The arrow between the gray circle on the left and the middle blue circle represents two types of errors, {\it Quantization error} and {\it Clipping error}, respectively, totally denoted as {\it QC errors}. The arrow between the middle and right circles represents {\it Temporal errors}. These errors are addressed in the two stages of FAS. More specifically, \textbf{\textit{Stage 1}} addresses {\it QC errors} through full-parameter fine-tuning. \textbf{\textit{Stage 2}} employs a layer-wise and neuron-wise coarse-to-fine calibration optimization strategy to minimize {\it Temporal errors}. The detailes of the algorithm are presented in \textbf{Appendix}~\ref{sec:alg_details}.


\subsection{\textit{Stage 1:} Eliminating QC Errors}

Inspired by Eq.~(\ref{eq9}), we select a continuous step function to replace the activation function in ANN to approximate the activation function of SNNs, thereby eliminating \textit{QC errors}. 
In this paper, we consider  Quantization Clip-Floor-Shift (QCFS) function~\cite{Bu2022OptimizedPI},
which is described as:
\begin{equation}
\label{eq10}
    a^l = f(W^l a^{l-1})
    = \text{clip}\bigl(\tfrac{\lambda^l}{L}
      \lfloor\tfrac{W^l a^{l-1}L}{\lambda^l}+\tfrac12\rfloor,\;0,\;\lambda^l\bigr),
\end{equation}
where \( \lambda^l \) is maximum value in ANN activation function, mapped to the thresholds \(\theta^l\) in SNN. \(L\) refers to the number of simulation steps. In contrast, existing methods, including QCFS, focus on eliminating \textit{QC errors} that need to be trained from scratch to get the initial ANN model~
\cite{bu2023optimal, wang2023toward, hao2023reducing}. However, it is time-consuming and impractical for LLMs. To address this, we novelty choose the pre-trained LLMs as the initial model that can effectively reduce the computational cost.

\subsection{Observations from Post-Stage-1 Analysis}

To better understand the source of the \textit{temporal error}, we conducted an in-depth analysis. 

\noindent \textbf{Definition.} We define \textit{Theoretical Maximum Spike Count} \(\psi ^i\) of neuron \(i\) in the \(l\)-th layer as max of \textit{Theoretical Spike Count} \(\tau _{theor}\)  during the interval \([0,T]\) for all data, that is:
\begin{equation}
    \psi ^i = \text{Max}(\tau _{theor})=\text{Max}( \tfrac{a^i T}{\lambda^i}),
\end{equation}t
where \(a^l/\lambda^l\) represents the normalized output in the ANN. \(\tau_{theor}\) denotes the number of spikes needed by an SNN neuron to accurately represent \(a^l\).  \(\psi^i\) indicates that the \(\tau_{theor}\) in the \(i\)-th neuron will not exceed \(\psi^i\). If the spike count in the SNN matches \(\tau_{theor}\), the conversion error would be zero. However, due to the \textit{temporal error}, the actual spike count typically does not equal \(\tau_{theor}\). Based on the definitions, 
we have made the following three observations:

\begin{figure*}[!ht]
    \centering
    \subfigure[O-1: `w/' and `w/o' denotes the case with and without optimization.]{
        \includegraphics[width=0.3\textwidth]{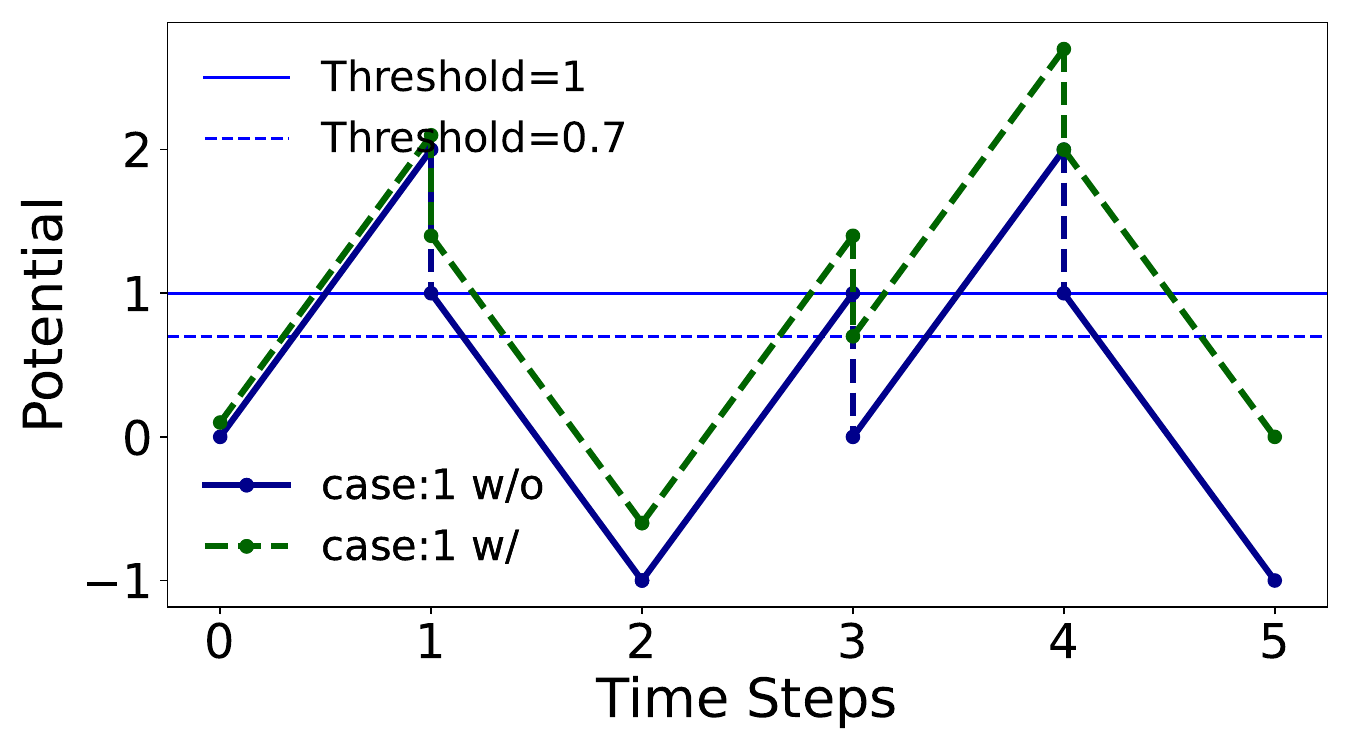}
        \label{case}
    }
    \subfigure[O-2: The distribution of theoretical spike counts with layers.]{
        \includegraphics[width=0.3\textwidth]{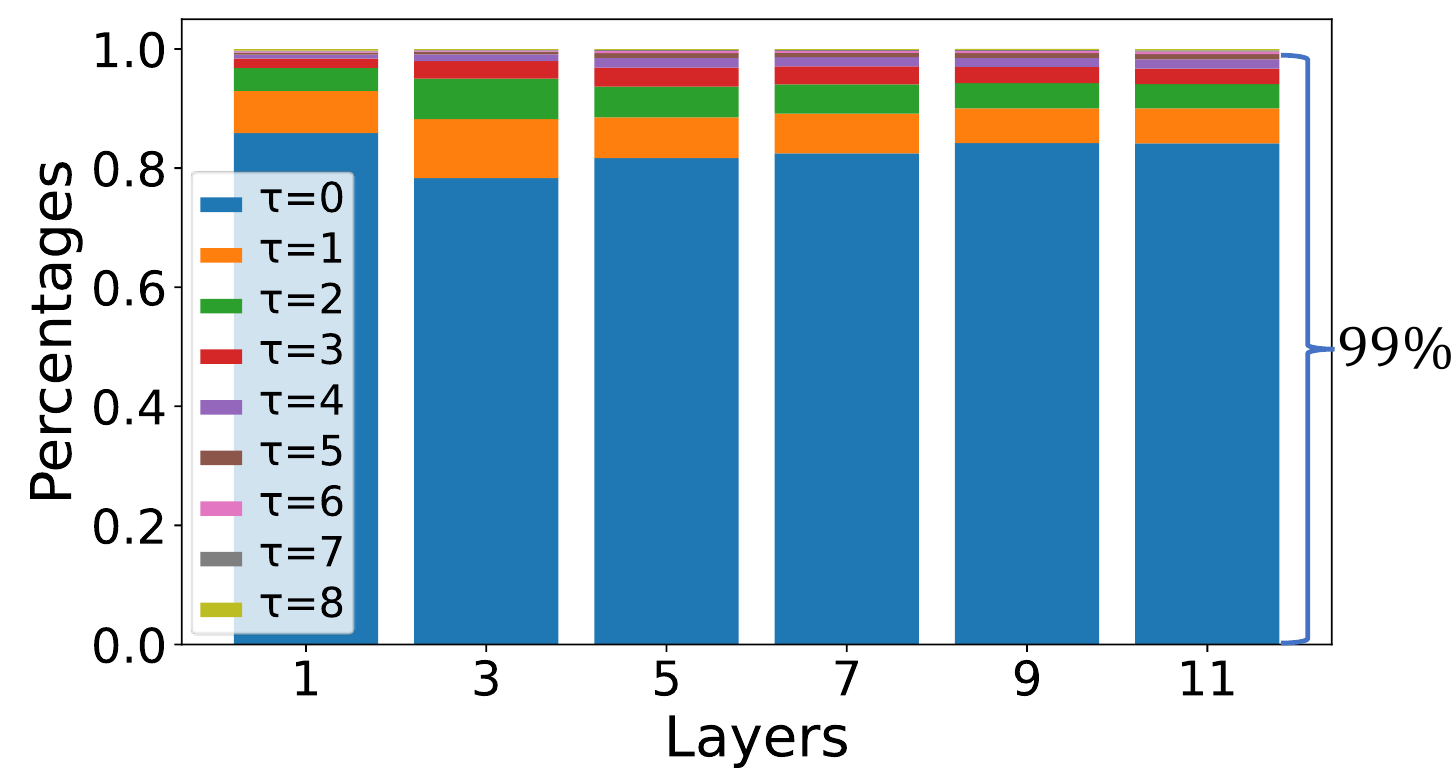}
        \label{TheoreticalSpike}
    }
    \subfigure[O-3: The distribution of theoretical maximum spike counts.]{
        \includegraphics[width=0.3\textwidth]{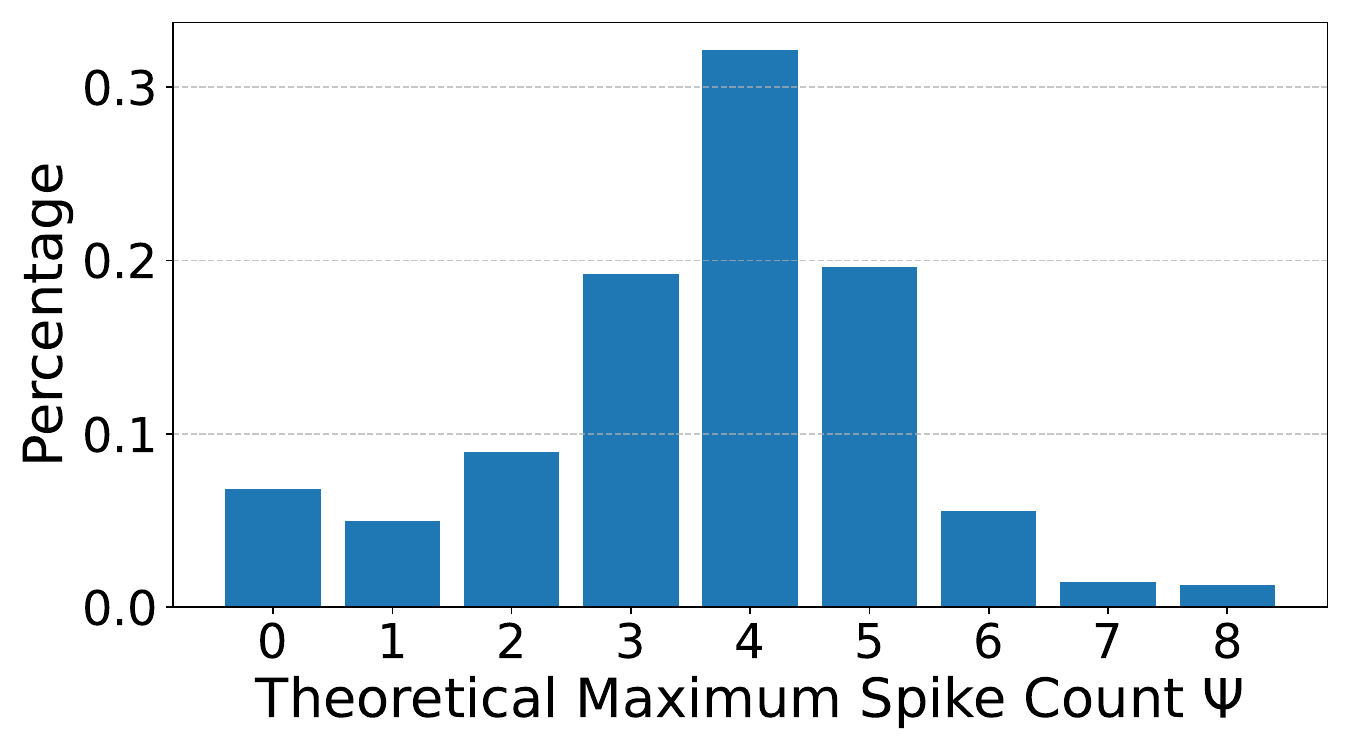}
        \label{MaximumSpike}
    }
    \caption{Illustration of our observations.}
    \label{fig:case}
\end{figure*}

\begin{tcolorbox}[colback=gray!20, colframe=gray!20, arc=1mm, breakable, left=1mm, right=1mm, top=0.1mm, bottom=0.1mm]
\textbf{Observation 1.} By lowering the thresholds when \(\tau _{theor} < T\), the \textit{temporal error} can be reduced.
\end{tcolorbox}

\noindent The maximum activation value \(\theta^l\)  in the SNN is aligned with the upper activation bound \(\lambda^l\) of the ANN, thereby eliminating clipping error.
However, when \(a^l\) is lower than \(\theta^l\), \(\theta^l\) can be set in the range \([a^l, \theta^l]\) without affecting the firing rate mapping to \(a^l\), provided that other data is not considered.
Moreover, \textit{temporal error}, caused by firing more or fewer spikes than expected, can be expressed as:
\begin{equation}
    \mathrm{Error}_{T}
= \tfrac{\bigl|\tau_{\mathrm{real}} \cdot \theta^l - \tau_{\mathrm{theor}} 
\cdot \lambda^l\bigr|}{T}
\label{eq:temporalError}
\end{equation}
where \( \tau_{real} \) represents the number of spikes in practice. As \(\theta^l\) decreases, the \textit{temporal error} decreases as well. Lowering the threshold \(\theta^l\) may slightly affect the firing rate representation, but this can be mitigated by optimizing initial membrane potentials. This effect is minor compared to the reduction in \textit{temporal error}. The following example illustrates this point.

Consider two pre-synaptic neurons in the \( (l-1) \)-th layer connected to a postsynaptic neuron in the \( l \)-th layer, as shown in Figure~\ref{case}, denoted as case 1. Note that we assume $\theta^l=1$, $\lambda^l=1$, $\tau_{theor}=2$, and $T=5$ in case 1. If the pre-synaptic neurons fire at \( t = 1, 3, 4 \) and \( t = 2, 5 \), respectively, the postsynaptic neuron will fire three spikes at \( t = 1, 3, 4 \). Thus, the number of the spikes, i.e., $\tau_{real}$, is three. Based on Eq.~(\ref{eq:temporalError}), the \textit{temporal error}, i.e., $\text{Error}_{T}$, is $0.2$. By lowering the SNN thresholds $\theta^l$ to $0.7$ and resetting the initial membrane potentials, the \textit{temporal error} is reduced to $0.02$.

\begin{tcolorbox}[colback=gray!20, colframe=gray!20, arc=1mm,  left=1mm, right=1mm, top=0.1mm, bottom=0.1mm]
\textbf{Observation 2.}  \(\tau_{\text{theor}} \le T/2\) in $99$\% of the cases in each layer, lowering the threshold in each layer can effectively decrease \textit{temporal error}.
\end{tcolorbox}

\noindent We further investigate the distribution of theoretical spike count \(\tau_{\text{theor}}\) in each layer.
Figure~\ref{TheoreticalSpike} shows the \(\tau_{\text{theor}}\) distribution for the converted GPT-2 model on WikiText-103 with 8 time steps. It is observed that \(\tau_{\text{theor}} \leq 4\) accounts for $99$\% of cases, while \(\tau_{\text{theor}} \geq 5\) accounts for only $1$\%. This indicates that neurons in the SNN typically need to fire at most 4 spikes to represent ANN activation values. The range of \(r^l\) is \([0, \tau_{theor}\theta^l/{T}] = [0, \theta/2]\). Based on this, we can lower the threshold to within \([\theta/2, \theta]\). Combined with \textbf{Observation 1}, this effectively reduces \textit{temporal error} for all neurons. For example, decreasing the threshold in each GPT-2 layer to $60$\% of its original value shows performance improvement.  Detailed analyses are in the ablation study.

\begin{tcolorbox}[colback=gray!20, colframe=gray!20, arc=1mm, breakable, left=1mm, right=1mm, top=0.1mm, bottom=0.1mm]
\textbf{Observation 3.} \(\psi^i > T/2\) also occupies a significant portion, necessitating threshold optimization for each neuron to achieve better performance.
\end{tcolorbox}
\noindent We investigate the distribution of the theoretical maximum spike count \(\psi^i\) for each neuron, excluding the top 1\% of \(\tau_{\text{theor}}\) cases. As shown in Figure~\ref{MaximumSpike}, \(\psi^i\) varies among neurons, with most being less than \(T\). This suggests that lowering the threshold, as \textbf{Observation 2}, effectively reduces \textit{temporal error}. However, unlike \textbf{Observation 2}, where \(\tau_{\text{theor}} > T/2\) was rare ($1$\%), many neurons have \(\psi^i > T/2\). Thus, reducing the threshold in each layer too much jeopardizes the performance of these neurons, while reducing it slightly will not sufficiently decrease the \textit{temporal error}. Thus, optimizing the threshold of each neuron is necessary.

In summary, our post-stage-1 study shows that lowering the threshold can reduce \textit{temporal error}. The second stage, which is detailed next, is designed based on this observation.

\subsection{\textit{Stage 2:} Eliminating Temporal Error}
To minimize \textit{temporal error}, we first reduce the overall \textit{temporal error} through {\it layer-wise calibration} of the thresholds and initial membrane potentials. Then, we perform {\it neuron-wise calibration} to optimize each neuron.

\textit{Layer-wise calibration (LWC)}: 
As shown in \textbf{Observation 2}, adjusting the threshold and initial membrane potentials in each layer can reduce \textit{temporal error}. Thus, we optimize the thresholds and initial membrane potentials as:
\begin{equation}
    \widehat{\theta}^l  = \alpha^l \ast \theta^l, \widehat{v}(0)^l = \beta^l \ast \widehat{\theta}^l,
\end{equation}
where \(\theta^l\) is the threshold of the \(l\)-th layer, \(v(0)^l\) is the initial membrane potential of the \(l\)-th layer; \(\alpha^l\) and \(\beta^l\) are their optimization weights; \(\widehat{\theta}^l\) and \(\widehat{v}(0)^l\) represent the optimized thresholds and initial membrane potentials, respectively. The optimal value of \(\alpha^l\) can be determined by analyzing the distribution of theoretical spike counts in the training set.

\textit{Neuron-wise calibration (NWC)}: For each neuron, we set a trainable threshold \(\theta^l_i\) and initial membrane potential \(v(0)^l_i\), with initial values set after layer-wise calibration. Using the ANN as a guide, we input the same data into both the ANN and SNN for forward propagation, minimizing the distance between the firing rate of each SNN neuron and the output of the corresponding ANN neuron. 
We freeze all model parameters except the thresholds and initial membrane potentials of the IF neurons, then update these parameters for each neuron to achieve neuron-wise optimization. Due to the discrete and non-differentiable nature of spikes, standard backpropagation cannot be used here, so we employ BPTT~\cite{Lee2016TrainingDS,Wu2017SpatioTemporalBF} for calibration.
Next, the proposed loss functions used for backpropagation are discussed, i.e., \textit{activation align loss} and \textit{logits loss}.

\textit{Activation align loss:} 
To minimize the \textit{temporal error} of SNN, the firing rate of IF neurons should align with the activation values of the ANN. The loss function can be described as:
\begin{equation}
    L^{al}_i=mse(a^l_i, r^l_i)=mse(f(W^{l}a^{l-1}), \tfrac{\sum_{t=1}^{\rho}s^{l}_i\theta ^{l}_i}{T}),
\end{equation}
where \(\rho\) denoted the time steps used for calibration. \(a^l_i\) is the output of the \(i\)-th neuron in the \(l\)-th layer, while \(r^l_i\) signifies the fire rate of the \(i\)-th neurons in the \(l\)-th layer.

\textit{Logits loss:} Following ~\cite{Hinton2015DistillingTK}, we use logits loss, which lets the SNN learn the prediction distribution of the ANN. To measure the distance between two distributions, we choose KL-divergence:
\begin{equation}
    L_{logits}=-{\textstyle \sum_{i}^{c}\text{Softmax}(\frac{a_i}{T})\text{log}(\text{Softmax}(\frac{r_i}{T})) },
\end{equation}
where \( T \) is the temperature parameter, \( a_i \) represents the output of the ANN, \( r_i \) represents the spike fire of the SNN, and \( c \) represents the number of classes. Therefore, the total loss contains two terms:
\begin{equation}
    L_{all}=\lambda _1  \textstyle\sum_{i}L^{al}_i + \lambda _2 L_{logits},
    \label{eq:L_all}
\end{equation}
where \(\lambda _1 \) and \(\lambda_2\) are the hyper-parameters that control the weight of activation align loss and logits loss, respectively. 
A more detailed implementation of \textit{NWC} is provided in the \textbf{Appendix}~\ref{sec:detallnwc}.
\section{Experiments}
\label{sec:experiments}

\subsection{Datasets \& Baselines \& Settings}
To evaluate  FAS , we selected various language and vision-language tasks (see \textbf{Appendix}~\ref{sec:dataset}). In addition, we chose various SOTA models, including LLM and multimodal LLM, as the peer competitors, and their details are provided in {\bf Appendix ~\ref{sec:baseline}}. 
Note that the experimental settings, more parameter studies, and more efficacy studies are presented in \textbf{Appendices}~\ref{sec:exp_setup},~\ref{sec:appendix_para_study}, and~\ref{sec:appendix_efficacy_study}, respectively.

\subsection{Experiments on NLU Tasks}
\label{sbsec:NLU_tasks}

\textbf{Performance Analysis on Bert:} Table~\ref{table1} compares FAS with other SOTA models on BERT, demonstrating FAS achieves new SOTA performance across seven text classification datasets. Specifically, on the QNLI and MRPC tasks, FAS surpasses other directly trained SNN models (the second block of Table~\ref{table1}) by at least $5$\%. Compared to other ANN-SNN methods (the third block of Table~\ref{table1}), FAS outperforms the QCFS by $10$\% on the RTE task. Notably, FAS achieves superior performances with time steps of only {\bf $4$}, indicating the low latency of the model. Moreover, experience with 8 time steps are provided in \textbf{Appendix}~\ref{sec:addExpBERT}. Additionally, we justify FAS can effectively approximate the GLUE function of BERT and performer well on different scale BERT (see \textbf{Appendices}~\ref{sec:appendix_app_gelu} and~\ref{sec:appendix_pre_train}).

\begin{table*}[!b]
\centering
\small
\caption{
Comparing FAS with SOTA models of BERT on the GLUE evaluation set. \textit{S} denotes whether an SNN or not. \textit{T} is the time steps. $^*$ denotes non-convergence. $^\dagger$ indicates additional time steps required to gather the necessary prior information. The three blocks group models of non-SNN, direct trained and ANN-SNN converted.
}
\resizebox{1.0\textwidth}{!}{
\begin{tabular}{llllllllll}
\toprule
\textbf{Model} & \textit{S} &\textit{T }&\textbf{QQP} & \textbf{MNLI-m} & \textbf{SST-2} & \textbf{QNLI} & \textbf{RTE} & \textbf{MRPC} & \textbf{STS-B} \\
\midrule
BERT \cite{Devlin2019BERTPO} & \ding{55} &N/A &90.71 & 83.91 & 92.32 & 90.66 & 65.70 & 84.07/88.85 &88.64/88.48 \\
CBoW \cite{Wang2018GLUEAM} &\ding{55} &N/A& 75.00 & 57.10 & 79.50 &  62.50 & 71.90 &  75.00/83.70 &70.60/71.10\\
BiLSTM \cite{Wang2018GLUEAM} & \ding{55}&N/A & 85.30 & 66.70 & 87.50 &  77.00 & 58.50 &  77.90/85.10 &71.60/72.00\\
BiLSTM + Attn, CoVe \cite{Wang2018GLUEAM}&\ding{55}  &N/A& 83.50 & 67.90  & 89.20 &  72.50 & 58.10 & 72.80/82.40 & 59.40/58.00\\
GenSen \cite{subramanian2018learning} &\ding{55} &N/A& 82.60 & 71.40  & 87.20 & 62.50 & 78.40 & 80.40/86.20 &81.30/81.80\\
\midrule
SNN-TextCNN \cite{Lv2023SpikingCN} &\ding{51} & 50 & \(0.00^\star \) & 64.91 & 80.91 & 64.91 & 47.29 & -/80.62 & \(0.00^\star \)/- \\
spikeBERT \cite{lv2024spikebertlanguagespikformerlearned} &\ding{51} &4 & 68.17 & 71.42 & 85.39 & 66.37 & 57.47 & -/81.98 & -/\(18.73^\star \) \\
SpikeLM \cite{xing2024spikelm} &\ding{51}& 4 & -&77.10&87.00&85.30&69.00&-/85.70&84.90/-   \\
SpikingBERT \cite{bal2024spikingbert} & \ding{51}&60 & 86.82 &  78.10 &  88.19 &  85.20 &  66.06 &  79.17/85.15 &  82.20/81.90\\
\midrule
SPR \cite{hao2023reducing} & \ding{51}&8 ($16^\dagger $) &87.48&77.56 & 90.48&87.75&64.98&78.68/85.76&86.71/86.50    \\ 
QCFS \cite{bu2023optimal}  & \ding{51}&8 &88.42&79.57&89.91&86.80&56.68&78.92/85.37&86.18/85.82\\ 
COS \cite{Hao2023BridgingTG}  & \ding{51}&8 ($8^\dagger$) &88.85&79.91&89.79&87.37&63.18&79.66/86.33&86.49/86.23\\ 
\midrule
\textbf{FAS (BERT)} & \ding{51} &\textbf{4}& \textbf{90.38}  & \textbf{82.77} & \textbf{91.17} & \textbf{90.13} & \textbf{66.06} & \textbf{86.02/90.22} & \textbf{87.46/87.26} \\
\bottomrule
\end{tabular}
}
\label{table1}
\end{table*}

\textbf{Energy Analysis on Bert:} As shown in Table~\ref{tab:gpt2_energy}, FAS can effectively reduce the energy consumption yet with the SOTA performance across all time steps. Note that the performance of FAS even exceeds the baseline BERT model with only 8.42\% energy consumption. In addition, compared to others, i.e., QCFS, SRP, and COS, FAS has fewer time steps under similar energy consumption. This indicates that FAS has a faster inference speed than others, especially when deploying on hardware. Note that the details of the energy consumption are described in \textbf{Appendix}~\ref{sec:energy_analysis}.

\begin{table*}[!t]
    \centering
    \small
    \caption{Energy efficiency analysis of FAS on the QQP task of BERT model. Also, SRP and COS need an additional 16 time steps to gather the necessary prior information. \textcolor{blue}{\( \uparrow\)} and \textcolor{red}{\( \downarrow\)} denote the performance is better or worse than the baseline BERT model, respectively.}
    \resizebox{1.0\textwidth}{!}{
    \begin{tabular}{lllllllll}
         \toprule
         \multirow{2}*{\textit{T}} & \multicolumn{2}{c}{\textbf{FAS}} & \multicolumn{2}{c}{\textbf{QCFS}} & \multicolumn{2}{c}{\textbf{SRP}} & \multicolumn{2}{c}{\textbf{COS}} \\ 
         ~ & \textbf{Accuracy} & \textbf{Energy (\%)} & \textbf{Accuracy} & \textbf{Energy (\%)} & \textbf{Accuracy} & \textbf{Energy (\%)}& \textbf{Accuracy} & \textbf{Energy (\%)} \\
         \midrule
         N/A (BERT) & 90.66 & 100 & 90.66 & 100 & 90.66 & 100 & 90.66 & 100 \\
         \midrule
         16 & \textbf{90.75} (\textcolor{blue}{\( \uparrow\)} 0.09) & 8.42 & 87.53 (\textcolor{red}{\( \downarrow\)} 3.13) & 2.50 & 87.28 (\textcolor{red}{\( \downarrow\)} 3.38) & 5.46 & 87.31 (\textcolor{red}{\( \downarrow\)} 3.35) & 4.98 \\
         8 & 90.20 (\textcolor{red}{\( \downarrow\)} 0.46) & 4.56 & 86.84 (\textcolor{red}{\( \downarrow\)} 3.82) & 1.28 & 87.41 (\textcolor{red}{\( \downarrow\)} 3.25) & 4.11 & 86.82 (\textcolor{red}{\( \downarrow\)} 3.84) & 3.77 \\
         4 & 90.38 (\textcolor{red}{\( \downarrow\)} 0.28) & 3.14 & 85.01 (\textcolor{red}{\( \downarrow\)} 5.56) & 0.65 & 87.15 (\textcolor{red}{\( \downarrow\)} 3.51) & 3.42 & 87.31 (\textcolor{red}{\( \downarrow\)} 3.35) & 3.16 \\
         2 & 89.69 (\textcolor{red}{\( \downarrow\)} 0.97) & 1.88 & 82.19 (\textcolor{red}{\( \downarrow\)} 8.47) & 0.32 & 86.23 (\textcolor{red}{\( \downarrow\)} 4.43) & 3.05 & 86.66 (\textcolor{red}{\( \downarrow\)} 4.00) & 2.84 \\
         1 & 88.94 (\textcolor{red}{\( \downarrow\)} 1.72) & 1.14 & 81.55 (\textcolor{red}{\( \downarrow\)} 9.11) & \textbf{0.12} & 84.94 (\textcolor{red}{\( \downarrow\)} 5.72) & 2.84 & 84.81 (\textcolor{red}{\( \downarrow\)} 5.85) & 0.31 \\
         \bottomrule
    \end{tabular}
    }
    \label{tab:gpt2_energy}
\end{table*}

\begin{table*}[!t]
\centering
\small
\caption{
Comparing the accuracy of zero-shot tasks between FAS and SOTA OPT models.
}
\resizebox{1.0\textwidth}{!}{
\begin{tabular}{lllllllllll}
\toprule
\textbf{Model} & \textit{S} &\textit{T } & \textbf{Energy (\%)} &\textbf{PIQA} & \textbf{ARC} & \textbf{OpenbookQA} & \textbf{Winogrande} & \textbf{COPA} & \textbf{WSC} & \textbf{RTE} \\
\midrule
OPT-7B \cite{zhang2022opt} & \ding{55} &N/A & 100 & 76.26 & 65.57 & 27.60 & 65.43 & 81.00 & 82.05 &55.25\\
\midrule
\multirow{3}*{\textbf{FAS (OPT-7B)}} & \ding{51}& \textbf{8} & 3.37&72.74&63.97 & 27.60&60.30&\textbf{84.00}&77.29&53.07    \\ 
~  & \ding{51}&16 & 5.06&73.23&\textbf{64.73}&27.00&\textbf{60.38}&83.00&\textbf{77.66}&\textbf{55.60}\\ 
~  & \ding{51}&32 & 8.41&\textbf{74.05}&64.60&\textbf{27.80}&60.06&82.00&77.29&55.23\\
\bottomrule
\end{tabular}
}
\label{tab:tableOPT}
\end{table*}

\textbf{Performance \& Energy Analysis on OPT-7B:} We also conducted the experiments on the larger model, i.e., OPT7B, to further justify the effectiveness of FAS. As presented in Table~\ref{tab:tableOPT}, we compared 
the performance of OPT and the spiking OPT converted by FAS under different time steps. The results show that FAS, as well as or outperforming the original OPT-7B model across different tasks. Specifically, in OpenBookQA and COPA, FAS under 8 time steps achieved the same or higher performance than OPT. In the RTE task, FAS with 16 time steps surpassed the performance of OPT. 
Additionally, FAS can significantly reduce energy consumption across the different time steps.

\subsection{Experiments on NLG Tasks}
Table~\ref{table2} presents the results of FAS using the GPT-2 architecture on the Enwik8 and WikiText-103 datasets. It outperforms all other methods with low latency inference (\textit{T}=16). 
For Enwik8, FAS achieves 0.968 BPB, whereas the QCFS and SRP methods reach 1.016 and 1.014 BPB at 32 time steps. COS reaches 1.01 BPB at 16 time steps, but requires additional time steps for prior information. 
Moreover, on WikiText-103, FAS achieves a perplexity (PPL) of 16.84, while SpikeGPT reaches 18.01 PPL at 1024 time steps. FAS also outperforms other ANN-SNN conversion methods (the third block of Table~\ref{table2}) by a large margin. Note that FAS also has a high energy efficiency on GPT-2 across different time steps, and the details are provided in \textbf{Appendix}~\ref{sec:energy_analysis}. 

\subsection{Experiments on Vision-Language Tasks}
To validate the generalization of FAS, we conducted experiments on the multimodal Pailgemma-3B and LLaVa-1.5-7B. As shown in Table~\ref{tab:pailgame3B}, FAS outperformed Paligemma-3B on the HallusionBench and BLINK benchmarks with 8 time steps. Moreover, despite having only 3B parameters, FAS exceeded the performance of several larger 7B models, including MiniGPT-4 and Qwen-VL-8B, across all tasks. Additionally, for LLaVa-1.5-7B, FAS achieved high accuracy, with performance only 0.8 lower than the original LLaVa-1.5-7B model, using only 8 time steps. Furthermore, FAS outperformed both MiniGPT-4 and Qwen-VL-8B on all tasks, and exceeded the performance of Paligemma-3B on the BLINK task. Additionally, the case studies are provided in \textbf{Appendix}~\ref{sec:appendix_case_studies}. Note that the performance of the fine-tuned ANN-based LLMs and converted spiking LLMs are also evaluated in \textbf{Appendix}~\ref{sec:appendix_fine_tuned_ann}.

\begin{table*}[!t]
\centering
\small
\caption{
Compare the performance of FAS and SOTA multimodal LLMs on  vision-language tasks.
}
\resizebox{1.0\textwidth}{!}{
\begin{tabular}{llllllll}
\toprule
\multirow{2}*{\textbf{Model}} & \multirow{2}*{\textit{S}} & \multirow{2}*{\textit{T}} & \multicolumn{3}{c}{\textbf{HallusionBench}} & \multicolumn{1}{c}{\textbf{BLINK}} & \multicolumn{1}{c}{\textbf{MMMU}} \\
~ & ~ & ~ & \textbf{Question Pair Acc.} & \textbf{Figure Acc.} & \textbf{Question Acc.}  & \textbf{Test} &  \textbf{Val} \\
\midrule
MiniGPT-4-v2-7B \cite{zhu2023minigpt} & \ding{55} & N/A  &8.79 & 10.12&35.78 & 34.6 & -  \\ 
Qwen-VL-8B \cite{bai2023qwen} & \ding{55} & N/A & 5.93 & 6.65 &39.15 & - & - \\
Claude 3 \cite{sonoda2024diagnostic} & \ding{55} & N/A & 21.76 & 28.61 & 56.86&44.1  & 50.2 \\
Paligemma-3B \cite{beyer2024paligemma} & \ding{55} & N/A & 22.63 & 21.96 & 51.84 & 38.29 &32.88  \\
LLaVA-1.5-7B \cite{liu2024visual} & \ding{55}& N/A &15.31  &20.87  &52.78 &41.22  &33.66  \\
\midrule
\textbf{FAS (Paligemma-3B)} & \textbf{\ding{51} }& \textbf{8} &\textbf{25.27} & \textbf{23.41} & \textbf{53.52} &\textbf{38.92} &\textbf{29.67}  \\ 
\textbf{FAS (LLaVA-1.5-7B)} & \textbf{\ding{51} }& \textbf{8} &\textbf{18.68} & \textbf{19.78} & \textbf{51.41} &\textbf{40.37} &\textbf{31.33}  \\ 
\bottomrule
\end{tabular}
}
\label{tab:pailgame3B}
\end{table*}

\begin{table*}[!t]
\centering
\small
\begin{minipage}{0.49\textwidth}
\centering
\caption{
Comparing FAS with SOTA GPT models on  NLG dataset. {\bf `En8'} stands for Enwik8, with BPB as the metric. {\bf `WT'} is WikiText-103 using perplexity. The lower the better for both metrics.}
\resizebox{\textwidth}{!}{
\begin{tabular}{lllll}
\toprule
\textbf{Model} & \textit{S} & \textit{T} &\textbf{En8} &\textbf{WT} \\
\midrule
GPT-2 \cite{radford2019language} & \ding{55} &N/A& 0.96 & 16.53 \\
Transformer-SSA \cite{hussain2023information} &\ding{55} &N/A& 1.02 & 16.91 \\
\midrule
AstroSNN \cite{Shen2023AstrocyteEnabledAI} & \ding{51} &  $-$ & 1.14 & 32.97 \\
spikeGPT \cite{zhu2023spikegpt} & \ding{51}&1024 & 1.26 & 18.01 \\ \hline
SPR \cite{hao2023reducing} & \ding{51} &32 ($16^\dagger$) &1.01 & 19.24 \\
QCFS \cite{bu2023optimal} & \ding{51} &32 &1.02 & 19.36 \\
COS \cite{Hao2023BridgingTG}& \ding{51} &16 ($16^\dagger$)&1.01 & 19.15 \\
\midrule
\textbf{FAS (GPT-2)} & \ding{51} &\textbf{16} &\textbf{0.97} & \textbf{16.84} \\
\bottomrule
\end{tabular}}
\label{table2}
\end{minipage}%
\hfill
\begin{minipage}{0.49\textwidth}
\centering
\caption{
Impact of the parameter \(\rho\) in GPT-2. Baseline refers to the SNN without Stage 2 optimization.}
\resizebox{\textwidth}{!}{
\begin{tabular}{llllll}
\toprule
\textit{T} & \(\rho=1\) & \(\rho=2\) & \(\rho=4\) & \(\rho=6\) & \(\rho=8\) \\
\midrule
N/A (GPT-2) & 16.39 & 16.39 & 16.39 & 16.39 & 16.39 \\
\midrule
1 & \textbf{23.06} & 24.79 & 30.01 & 34.37 & 39.30 \\
2 & 20.79 & \textbf{19.30} & 19.84 & 20.79 & 21.98 \\
4 & 19.75 & 18.20 & \textbf{17.68} & 17.78 & 18.01 \\
6 & 19.42 & 17.91 & 17.32 & \textbf{17.23} & 17.29 \\
8 & 19.30 & 17.79 & 17.21 & \textbf{17.05} & 19.03 \\
16 & 19.17 & 17.67 & 17.09 & 16.93 & \textbf{16.84} \\
32 & 19.13 & 17.64 & 17.09 & 16.94 & \textbf{16.83} \\
\bottomrule
\end{tabular}}
\label{table3}
\end{minipage}
\end{table*}

\subsection{Ablation Study and Impact of Hyper-Parameters}
\label{sec:ablation}

\paragraph{Parameter \(\rho\) :}
\begin{wrapfigure}{r}{0.5\columnwidth}
  \centering
  \begin{minipage}[b]{0.245\columnwidth}
    \centering
    \includegraphics[width=\linewidth]{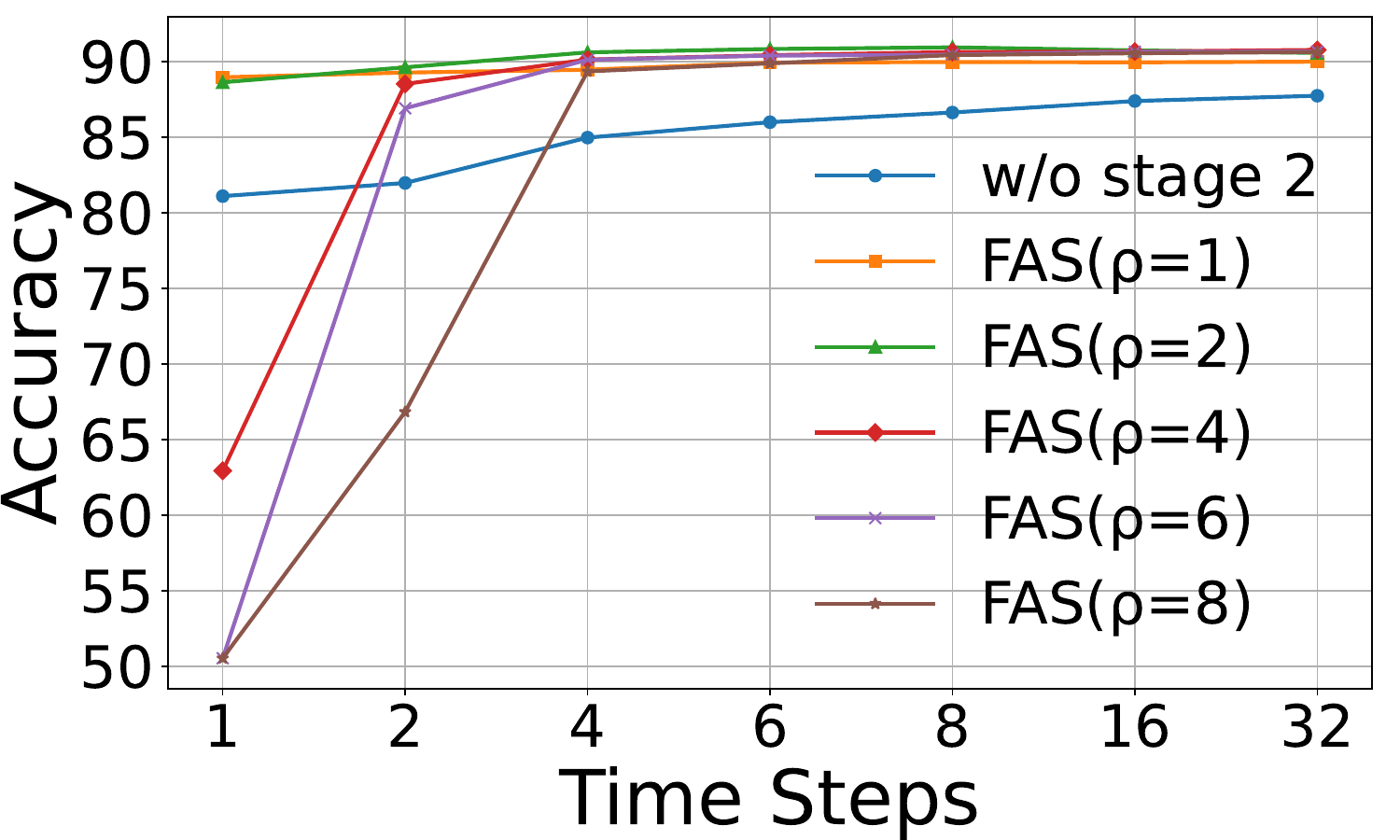}
    \caption*{(a) BERT on QNLI.}
  \end{minipage}%
  \begin{minipage}[b]{0.245\columnwidth}
    \centering
    \includegraphics[width=\linewidth]{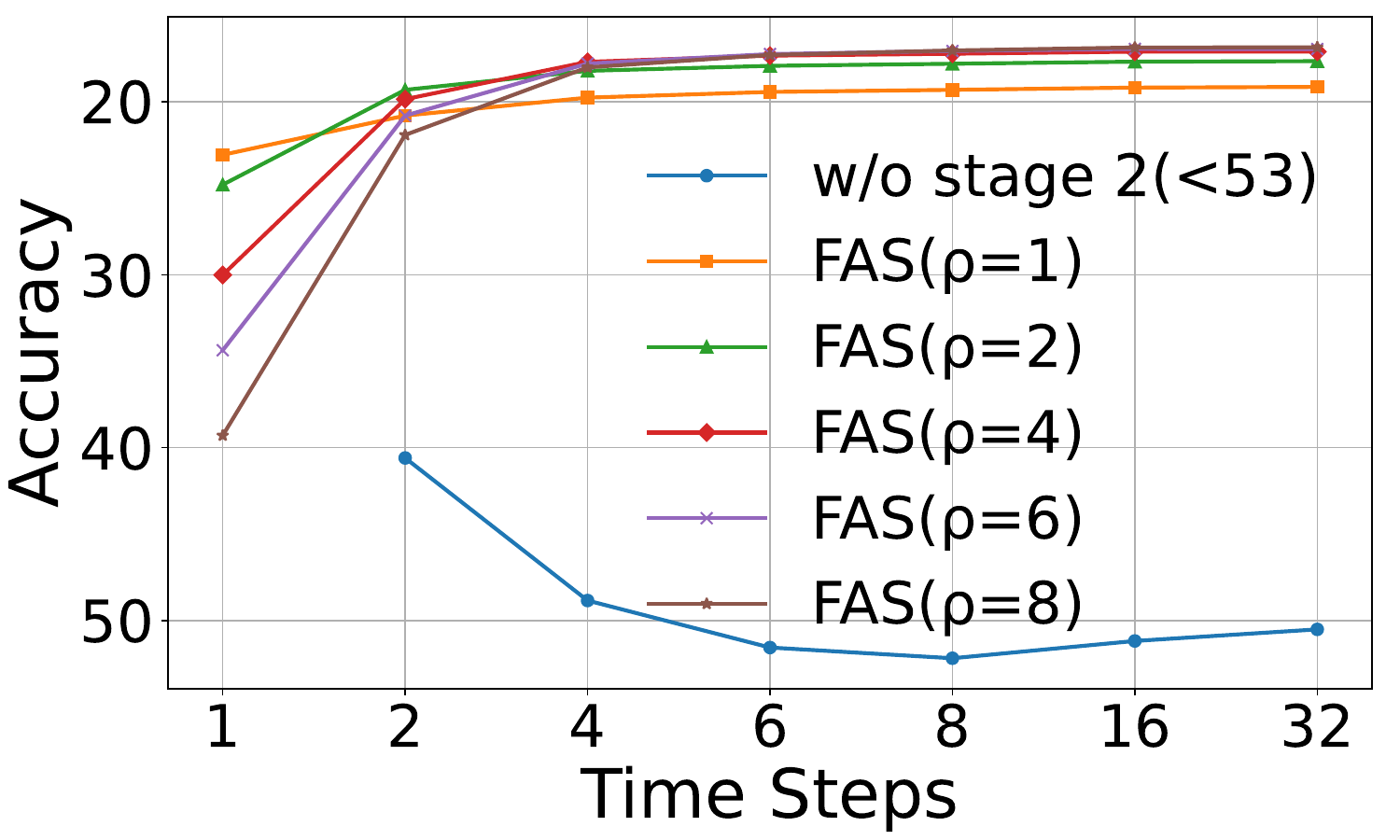}
    \caption*{(b) GPT-2 on WikiText.}
  \end{minipage}
  \caption{Impact of parameter \(\rho\)}
  \label{effect1}
\end{wrapfigure}
Figure~\ref{effect1} illustrates the performance of BERT and GPT-2 across different \(\rho\) values, revealing 
that \(\rho\) significantly affects performance. Table~\ref{table3} provides detailed performance for GPT-2
with various \(\rho\) values, showing that the SNN accuracy tends to converge as \(\rho\) gradually approaches \(T\). When \(\rho = T\), the best results are achieved. Furthermore, when \(T\) is very small $T \le  2$, setting \(\rho = T\) leads to significant improvements. More parameter studies on BERT are presented in \textbf{Appendix}~\ref{sec:param_rho_bert}.


\paragraph{LWC and NWC of Stage 2 :}
As shown in Table~\ref{stage1stage2}, compared to the first row, where none is present, \textit{NWC} and \textit{NWC} can both bring improvement.  The performance is the best for every time step when both are present. This indicates \textit{LWC} and \textit{NWC} have a positive impact on the performance of FAS. Moreover, we also evaluate the global performance of \textit{NWC} (see \textbf{Appendix}~
\ref{sec:appendix_efficacy_on_nwc}).

\paragraph{Parameter \(L\) :}

As shown in Table~\ref{ComparisonL}, increasing \(L\) initially can improve the performance of SNN, but causes a performance drop when \(L\) reaches 16. Choosing \(L\) is a trade-off between achieving high accuracy and maintaining low latency SNN.

\begin{figure*}[!t]
    \centering
    \subfigure[The distribution of thresholds vs. original threshold over varied  layers.]{
        \includegraphics[width=0.31\textwidth]{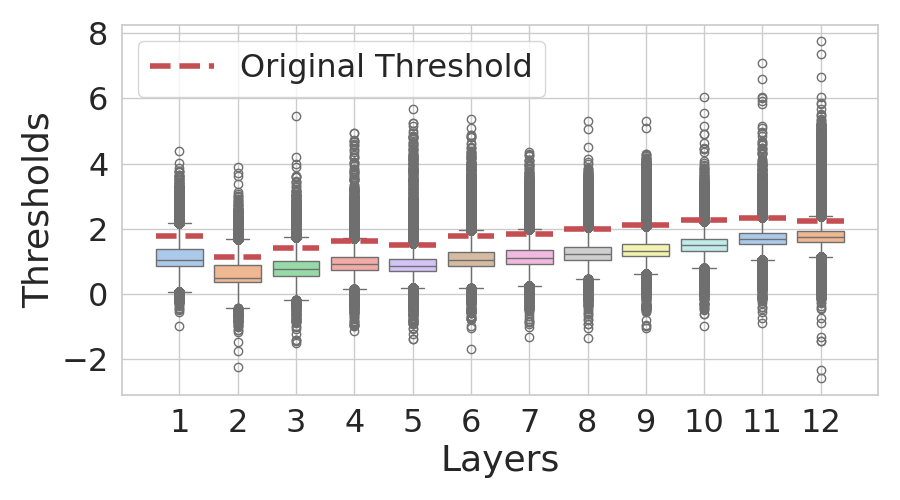}
        \label{fig:thr}
    }
    \subfigure[The distribution of initial membrane potentials over layers.]{
        \includegraphics[width=0.31\textwidth]{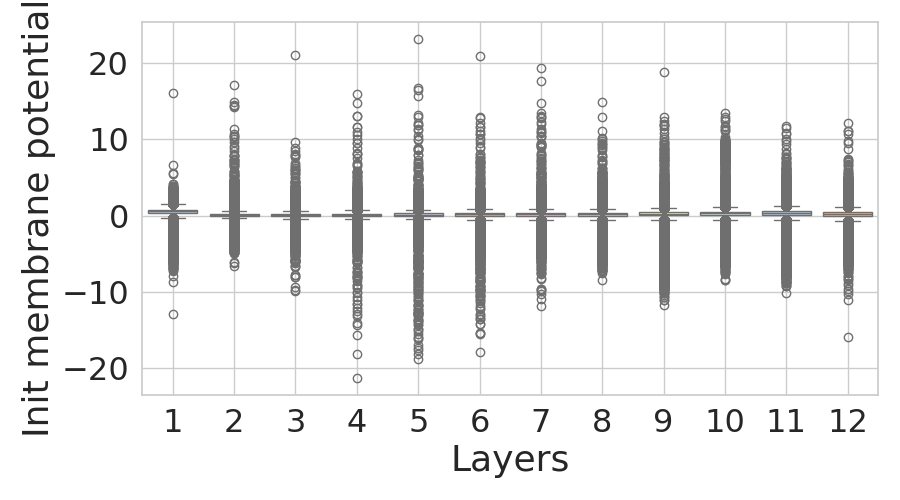}
        \label{fig:men}
    }
    \subfigure[Temporal error before and after the optimization over layers.]{
        \includegraphics[width=0.31\textwidth]{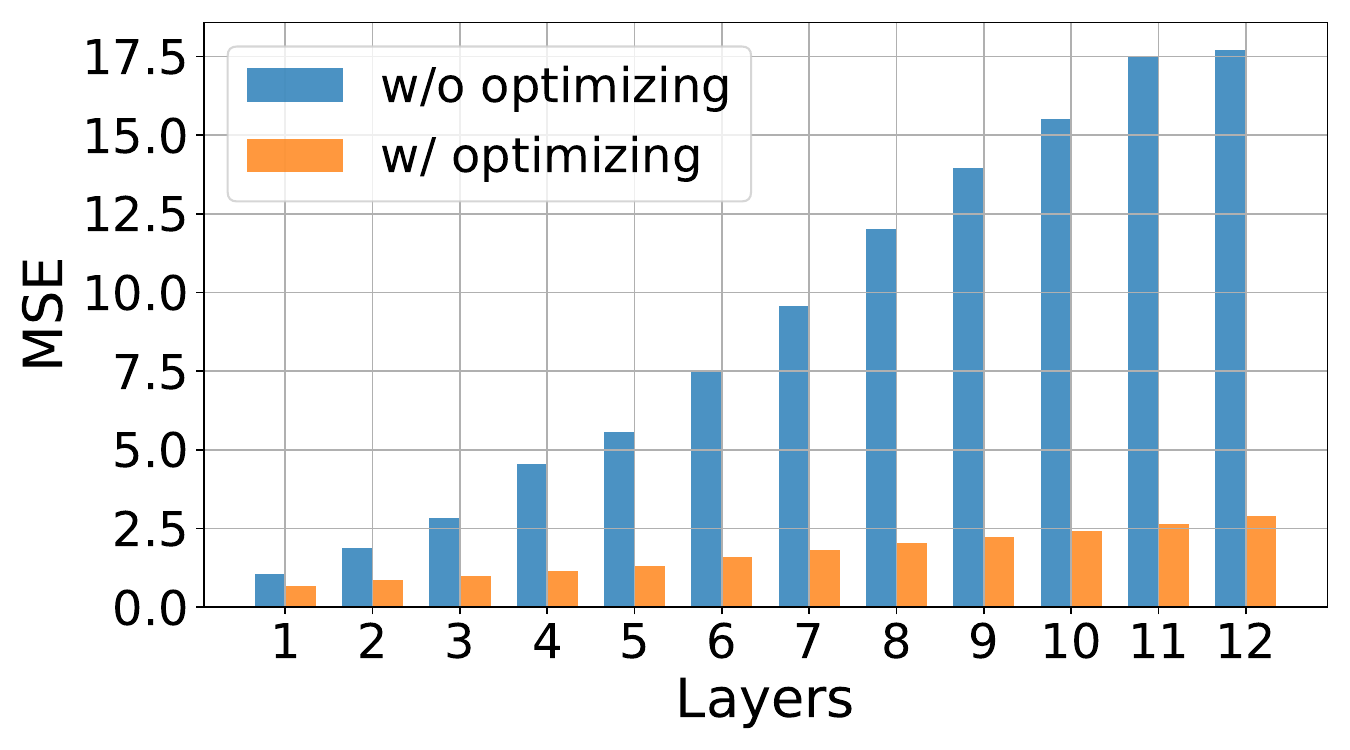}
        \label{fig:galaxy}
    }
    \caption{The effectiveness of FAS for threshold and initial membrane potentials optimization.}
    \label{fig:Activate}
\end{figure*}

\begin{table*}[!t] 
    \centering
    \begin{minipage}{0.48\textwidth} 
        \centering
        \small
        \caption{Ablation studies on \textit{LWC} and \textit{NWC}.}
        \resizebox{\linewidth}{!}{ 
        \begin{tabular}{lllllll}
            \toprule
            \textit{LWC} & \textit{NWC}&\textit{T=1}&\textit{T=2}&\textit{T=4}&\textit{T=8}&\textit{T=16} \\  \midrule
            \ding{55} &\ding{55} &66.41 &40.60 &48.85 & 52.19&51.59  \\ 
            \ding{51} &\ding{55}&64.78&28.99 & 23.96 & 22.81 & 20.22 \\ 
            \ding{55} &\ding{51}&40.01& 23.41& 18.30 & 17.10 &16.90  \\ 
            \ding{51} &\ding{51}&\textbf{39.30}& \textbf{21.98}& \textbf{18.01} & \textbf{17.03} &\textbf{16.84}  \\  \bottomrule
        \end{tabular}
        }
        \label{stage1stage2}
    \end{minipage}\hfill 
    \begin{minipage}{0.48\textwidth} 
        \centering
        \small
        \caption{Comparison on different $L$ values.}
        \resizebox{\linewidth}{!}{ 
        \begin{tabular}{lllllll}
            \toprule
            \textbf{L} & \textbf{ANN}&\textit{T}=1&\textit{T}=2&\textit{T}=4&\textit{T}=8&\textit{T}=16 \\  \midrule
            2 &17.89 &\textbf{20.28} &\textbf{18.50} &17.72 & 17.45&17.41  \\ 
            4 &16.87 &21.33 &18.68 &\textbf{17.51} & 17.10&16.99  \\ 
            8 &16.39&23.06&19.30 & 17.68 & \textbf{17.03} & \textbf{16.84} \\ 
            16 &16.17& 25.29 & 20.64 & 18.43 & 17.11 &16.90  \\ \bottomrule
        \end{tabular}
        }
        \label{ComparisonL}
    \end{minipage}
    \vskip -0.1in 
\end{table*}

\begin{wraptable}{r}{0.48\columnwidth}
  \centering
  \small
  \caption{Comparison of different \(\lambda _1, \lambda_2\) values.}
  \resizebox{0.48\columnwidth}{!}{
    \begin{tabular}{llllll}
        \toprule
        \(\mathbf{\lambda_1 : \lambda_2}\) & \textit{T}=1 & \textit{T}=2 & \textit{T}=4 & \textit{T}=8 & \textit{T}=16 \\ \midrule
        1:1 &\textbf{22.80} &19.12 &\textbf{17.58}  &17.14& 16.94  \\
        1:0 &47.06 &23.80 & 18.43 &17.36  &  17.10\\
        0:1 &22.80 & \textbf{19.12} & 17.58 &17.04& 16.89 \\
        1:0.0012 &23.06&19.30 & 17.68 & \textbf{17.02} & \textbf{16.85}  \\ \bottomrule
    \end{tabular}}
  \label{lenda}
\end{wraptable}
\paragraph{Parameter \(\lambda _1, \lambda_2\) :}
The impact of the weights \(\lambda_1\), for activation alignment loss, and \(\lambda_2\), for logits loss (see Eq.\ref{eq:L_all}) are shown in Table~\ref{lenda}.  They do show an impact on accuracy, especially with small time steps.  When \textit{T}$>$8, their impact becomes minor. More parameter studies on vision-language tasks and different \(\lambda _1: \lambda_2\) pairs are provided in \textbf{Appendix}~\ref{sec:appendix_lambda_vllm} and~\ref{sec:appendix_lambda_diff}, respectively.

\subsection{Effectiveness on Temporal Error}

The distributions of thresholds and initial membrane potentials before and after applying \textbf{\textit{Stage 2}} are shown in Figure~\ref{fig:Activate}a.  Most optimized thresholds are lower than their original, with a few outliers. Figure~\ref{fig:Activate}b depicts the optimized initial membrane potentials, primarily clustered above 0, with some values exceeding \(\pm 10\).
We further analyzed the MSE between \(r^l_i\) and \(a^l\), as shown in Figure~\ref{fig:Activate}c.  The MSE reductions in the last four layers are dramatic, \(83.95\%\), \(84.42\%\), \(84.8\%\), and \(86.66\%\), confirming that the proposed adjustments can effectively minimize temporal error.

\subsection{Effectiveness on LLM components}

FAS also can easily leverage existing techniques to achieve spiking Softmax and LayerNorm. For example, Jiang et al.~\cite{jiang2024spatio} introduced Universal Group Operators (UGO) to approximate non-linear functions using two fully connected layers with a ReLU activation. When using UGO on Softmax and LayerNorm, FAS is only to replace ReLU with IF neurons, enabling an efficient SNN conversion. As shown in Table~\ref{table:UGOBERT}, FAS (UGO BERT) nearly matches the performance of the original UGO BERT, while the QCFS method experiences noticeable drops (e.g., QQP: 84.68). These results confirm the effectiveness of FAS in incorporating spiking Softmax and LayerNorm during SNN conversion.


\begin{table}[ht]
\centering
\caption{Performance comparison between UGO-BERT and the converted SNN model.}
\label{table:UGOBERT}
\resizebox{0.75\textwidth}{!}{
\begin{tabular}{lccccc}
\toprule
                & QQP    &  QNLI     & RTE       &MRPC                &STS-B \\ \midrule
UGO BERT (ANN)        & 90.37  &   90.35 &   65.42    &  84.55/89.48   & 87.97/87.55    \\ 
\midrule
QCFS (UGO BERT)  & 84.68  &   83.41 &   55.89    &  76.98/83.54   & 82.32/81.77    \\ 
FAS (UGO BERT)   & \textbf{89.54}  &   \textbf{89.11} &   \textbf{64.38}    &  \textbf{83.58/88.51}   & \textbf{86.96/86.66}    \\ 
\bottomrule
\end{tabular}
}
\end{table}

\section{Conclusion}

FAS, a fast ANN-SNN conversion method tailored for LLMs, is presented.  It aims to leverage the low-cost computing benefits of Spiking LLMs while maintaining high performance. The conversion process is optimized through a two-stage strategy: firstly, full-parameter fine-tuning is applied so training from scratch is not needed.  Secondly, a coarse-to-fine calibration strategy is proposed to further minimize conversion errors, particularly temporal errors.  Experiments demonstrate that FAS can achieve both high performance and low latency.  The evaluation using both language and vision-language tasks with five LLMs shows that FAS can outperform SOTA approaches, maintaining accuracy comparable to ANN-based models, yet with low time steps.  

\bibliographystyle{plain}
\bibliography{neurips_2025}

\begin{thebibliography}{10}

\bibitem{bai2023qwen}
Jinze Bai, Shuai Bai, Shusheng Yang, Shijie Wang, Sinan Tan, Peng Wang, Junyang Lin, Chang Zhou, and Jingren Zhou.
\newblock Qwen-vl: A versatile vision-language model for understanding, localization, text reading, and beyond.
\newblock {\em arXiv preprint arXiv:2308.12966}, 1(2):3, 2023.

\bibitem{Bal2023SpikingBERTDB}
Malyaban Bal and Abhronil Sengupta.
\newblock Spikingbert: Distilling bert to train spiking language models using implicit differentiation.
\newblock In {\em AAAI Conference on Artificial Intelligence}, 2023.

\bibitem{bal2024spikingbert}
Malyaban Bal and Abhronil Sengupta.
\newblock Spikingbert: Distilling bert to train spiking language models using implicit differentiation.
\newblock In {\em Proceedings of the AAAI conference on artificial intelligence}, volume~38, pages 10998--11006, 2024.

\bibitem{beyer2024paligemma}
Lucas Beyer, Andreas Steiner, Andr{\'e}~Susano Pinto, Alexander Kolesnikov, Xiao Wang, Daniel Salz, Maxim Neumann, Ibrahim Alabdulmohsin, Michael Tschannen, Emanuele Bugliarello, et~al.
\newblock Paligemma: A versatile 3b vlm for transfer.
\newblock {\em arXiv preprint arXiv:2407.07726}, 2024.

\bibitem{brown2020language}
Tom Brown, Benjamin Mann, Nick Ryder, Melanie Subbiah, Jared~D Kaplan, Prafulla Dhariwal, Arvind Neelakantan, Pranav Shyam, Girish Sastry, Amanda Askell, et~al.
\newblock Language models are few-shot learners.
\newblock {\em Advances in neural information processing systems}, 33:1877--1901, 2020.

\bibitem{Bu2022OptimizedPI}
Tong Bu, Jianhao Ding, Zhaofei Yu, and Tiejun Huang.
\newblock Optimized potential initialization for low-latency spiking neural networks.
\newblock {\em ArXiv}, abs/2202.01440, 2022.

\bibitem{bu2023optimal}
Tong Bu, Wei Fang, Jianhao Ding, PengLin Dai, Zhaofei Yu, and Tiejun Huang.
\newblock Optimal ann-snn conversion for high-accuracy and ultra-low-latency spiking neural networks.
\newblock {\em arXiv preprint arXiv:2303.04347}, 2023.

\bibitem{cao2015spiking}
Yongqiang Cao, Yang Chen, and Deepak Khosla.
\newblock Spiking deep convolutional neural networks for energy-efficient object recognition.
\newblock {\em International Journal of Computer Vision}, 113:54--66, 2015.

\bibitem{davies2018loihi}
Mike Davies, Narayan Srinivasa, Tsung-Han Lin, Gautham Chinya, Yongqiang Cao, Sri~Harsha Choday, Georgios Dimou, Prasad Joshi, Nabil Imam, Shweta Jain, et~al.
\newblock Loihi: A neuromorphic manycore processor with on-chip learning.
\newblock {\em Ieee Micro}, 38(1):82--99, 2018.

\bibitem{deVries2023TheGE}
Alex de~Vries.
\newblock The growing energy footprint of artificial intelligence.
\newblock {\em Joule}, 2023.

\bibitem{Deng2021OptimalCO}
Shi-Wee Deng and Shi Gu.
\newblock Optimal conversion of conventional artificial neural networks to spiking neural networks.
\newblock {\em ArXiv}, abs/2103.00476, 2021.

\bibitem{Devlin2019BERTPO}
Jacob Devlin, Ming-Wei Chang, Kenton Lee, and Kristina Toutanova.
\newblock Bert: Pre-training of deep bidirectional transformers for language understanding.
\newblock In {\em North American Chapter of the Association for Computational Linguistics}, 2019.

\bibitem{Diehl2015FastclassifyingHS}
Peter~Udo Diehl, Daniel Neil, Jonathan Binas, Matthew Cook, Shih-Chii Liu, and Michael Pfeiffer.
\newblock Fast-classifying, high-accuracy spiking deep networks through weight and threshold balancing.
\newblock {\em 2015 International Joint Conference on Neural Networks (IJCNN)}, pages 1--8, 2015.

\bibitem{dubey2024llama}
Abhimanyu Dubey, Abhinav Jauhri, Abhinav Pandey, Abhishek Kadian, Ahmad Al-Dahle, Aiesha Letman, Akhil Mathur, Alan Schelten, Amy Yang, Angela Fan, et~al.
\newblock The llama 3 herd of models.
\newblock {\em arXiv preprint arXiv:2407.21783}, 2024.

\bibitem{Han2020RMPSNNRM}
{Han} et~al.
\newblock Rmp-snn: Residual membrane potential neuron for enabling deeper high-accuracy and low-latency spiking neural network.
\newblock {\em 2020 IEEE/CVF Conference on Computer Vision and Pattern Recognition (CVPR)}, pages 13555--13564, 2020.

\bibitem{han2020deep}
Bing Han and Kaushik Roy.
\newblock Deep spiking neural network: Energy efficiency through time based coding.
\newblock In {\em European conference on computer vision}, pages 388--404. Springer, 2020.

\bibitem{hao2023reducing}
Zecheng Hao, Tong Bu, Jianhao Ding, Tiejun Huang, and Zhaofei Yu.
\newblock Reducing ann-snn conversion error through residual membrane potential.
\newblock In {\em Proceedings of the AAAI Conference on Artificial Intelligence}, volume~37, pages 11--21, 2023.

\bibitem{Hao2023BridgingTG}
Zecheng Hao, Jianhao Ding, Tong Bu, Tiejun Huang, and Zhaofei Yu.
\newblock Bridging the gap between anns and snns by calibrating offset spikes.
\newblock {\em ArXiv}, abs/2302.10685, 2023.

\bibitem{Hinton2015DistillingTK}
Geoffrey~E. Hinton, Oriol Vinyals, and Jeffrey Dean.
\newblock Distilling the knowledge in a neural network.
\newblock {\em ArXiv}, abs/1503.02531, 2015.

\bibitem{hussain2023information}
Md~Shamim Hussain.
\newblock The information pathways hypothesis: Transformers are dynamic self-ensembles.
\newblock In {\em Proceedings of the 29th ACM SIGKDD Conference on Knowledge Discovery and Data Mining}, pages 810--821, 2023.

\bibitem{jiang2024spatio}
Yizhou Jiang, Kunlin Hu, Tianren Zhang, Haichuan Gao, Yuqian Liu, Ying Fang, and Feng Chen.
\newblock Spatio-temporal approximation: A training-free snn conversion for transformers.
\newblock In {\em The Twelfth International Conference on Learning Representations}, 2024.

\bibitem{Lee2016TrainingDS}
Junhaeng Lee, Tobi Delbr{\"u}ck, and Michael Pfeiffer.
\newblock Training deep spiking neural networks using backpropagation.
\newblock {\em Frontiers in Neuroscience}, 10, 2016.

\bibitem{Li2022EfficientAA}
Yang Li and Yi~Zeng.
\newblock Efficient and accurate conversion of spiking neural network with burst spikes.
\newblock In {\em International Joint Conference on Artificial Intelligence}, 2022.

\bibitem{Li2021AFL}
Yuhang Li, Shi-Wee Deng, Xin Dong, Ruihao Gong, and Shi Gu.
\newblock A free lunch from ann: Towards efficient, accurate spiking neural networks calibration.
\newblock {\em ArXiv}, abs/2106.06984, 2021.

\bibitem{li2024error}
Yuhang Li, Shikuang Deng, Xin Dong, and Shi Gu.
\newblock Error-aware conversion from ann to snn via post-training parameter calibration.
\newblock {\em International Journal of Computer Vision}, pages 1--24, 2024.

\bibitem{lian2023learnable}
Shuang Lian, Jiangrong Shen, Qianhui Liu, Ziming Wang, Rui Yan, and Huajin Tang.
\newblock Learnable surrogate gradient for direct training spiking neural networks.
\newblock In {\em IJCAI}, pages 3002--3010, 2023.

\bibitem{liu2024visual}
Haotian Liu, Chunyuan Li, Qingyang Wu, and Yong~Jae Lee.
\newblock Visual instruction tuning.
\newblock {\em Advances in neural information processing systems}, 36, 2024.

\bibitem{lozhkov2024fineweb-edu}
Anton Lozhkov, Loubna Ben~Allal, Leandro von Werra, and Thomas Wolf.
\newblock Fineweb-edu, May 2024.

\bibitem{lv2024spikebertlanguagespikformerlearned}
Changze Lv, Tianlong Li, Jianhan Xu, Chenxi Gu, Zixuan Ling, Cenyuan Zhang, Xiaoqing Zheng, and Xuanjing Huang.
\newblock Spikebert: A language spikformer learned from bert with knowledge distillation, 2024.

\bibitem{Lv2023SpikingCN}
Changze Lv, Jianhan Xu, and Xiaoqing Zheng.
\newblock Spiking convolutional neural networks for text classification.
\newblock In {\em International Conference on Learning Representations}, 2023.

\bibitem{merolla2014million}
Paul~A Merolla, John~V Arthur, Rodrigo Alvarez-Icaza, Andrew~S Cassidy, Jun Sawada, Filipp Akopyan, Bryan~L Jackson, Nabil Imam, Chen Guo, Yutaka Nakamura, et~al.
\newblock A million spiking-neuron integrated circuit with a scalable communication network and interface.
\newblock {\em Science}, 345(6197):668--673, 2014.

\bibitem{8891809}
Emre~O. Neftci, Hesham Mostafa, and Friedemann Zenke.
\newblock Surrogate gradient learning in spiking neural networks: Bringing the power of gradient-based optimization to spiking neural networks.
\newblock {\em IEEE Signal Processing Magazine}, 36(6):51--63, 2019.

\bibitem{radford2019language}
Alec Radford, Jeffrey Wu, Rewon Child, David Luan, Dario Amodei, Ilya Sutskever, et~al.
\newblock Language models are unsupervised multitask learners.
\newblock {\em OpenAI blog}, 1(8):9, 2019.

\bibitem{rueckauer2016theory}
Bodo Rueckauer, Iulia-Alexandra Lungu, Yuhuang Hu, and Michael Pfeiffer.
\newblock Theory and tools for the conversion of analog to spiking convolutional neural networks.
\newblock {\em arXiv preprint arXiv:1612.04052}, 2016.

\bibitem{rueckauer2017conversion}
Bodo Rueckauer, Iulia-Alexandra Lungu, Yuhuang Hu, Michael Pfeiffer, and Shih-Chii Liu.
\newblock Conversion of continuous-valued deep networks to efficient event-driven networks for image classification.
\newblock {\em Frontiers in neuroscience}, 11:294078, 2017.

\bibitem{Sengupta2018GoingDI}
Abhronil Sengupta, Yuting Ye, Robert~Y. Wang, Chiao Liu, and Kaushik Roy.
\newblock Going deeper in spiking neural networks: Vgg and residual architectures.
\newblock {\em Frontiers in Neuroscience}, 13, 2018.

\bibitem{Shen2023AstrocyteEnabledAI}
Guobin Shen, Dongcheng Zhao, Yiting Dong, Yang Li, Jindong Li, Kang Sun, and Yi~Zeng.
\newblock Astrocyte-enabled advancements in spiking neural networks for large language modeling.
\newblock {\em ArXiv}, abs/2312.07625, 2023.

\bibitem{song2024one}
Xiaotian Song, Andy Song, Rong Xiao, and Yanan Sun.
\newblock One-step spiking transformer with a linear complexity.
\newblock In {\em Proceedings of the Thirty-Third International Joint Conference on Artificial Intelligence}, pages 3142--3150, 2024.

\bibitem{sonoda2024diagnostic}
Yuki Sonoda, Ryo Kurokawa, Yuta Nakamura, Jun Kanzawa, Mariko Kurokawa, Yuji Ohizumi, Wataru Gonoi, and Osamu Abe.
\newblock Diagnostic performances of gpt-4o, claude 3 opus, and gemini 1.5 pro in “diagnosis please” cases.
\newblock {\em Japanese journal of radiology}, pages 1--5, 2024.

\bibitem{subramanian2018learning}
Sandeep Subramanian, Adam Trischler, Yoshua Bengio, and Christopher~J Pal.
\newblock Learning general purpose distributed sentence representations via large scale multi-task learning.
\newblock In {\em International Conference on Learning Representations}, 2018.

\bibitem{Wang2018GLUEAM}
Alex Wang, Amanpreet Singh, Julian Michael, Felix Hill, Omer Levy, and Samuel~R. Bowman.
\newblock Glue: A multi-task benchmark and analysis platform for natural language understanding.
\newblock In {\em BlackboxNLP@EMNLP}, 2018.

\bibitem{wang2023toward}
Ziming Wang, Yuhao Zhang, Shuang Lian, Xiaoxin Cui, Rui Yan, and Huajin Tang.
\newblock Toward high-accuracy and low-latency spiking neural networks with two-stage optimization.
\newblock {\em IEEE Transactions on Neural Networks and Learning Systems}, 2023.

\bibitem{Wu2017SpatioTemporalBF}
Yujie Wu, Lei Deng, Guoqi Li, Jun Zhu, and Luping Shi.
\newblock Spatio-temporal backpropagation for training high-performance spiking neural networks.
\newblock {\em Frontiers in Neuroscience}, 12, 2017.

\bibitem{xing2024spikelm}
Xingrun Xing, Zheng Zhang, Ziyi Ni, Shitao Xiao, Yiming Ju, Siqi Fan, Yequan Wang, Jiajun Zhang, and Guoqi Li.
\newblock Spikelm: Towards general spike-driven language modeling via elastic bi-spiking mechanisms.
\newblock {\em arXiv preprint arXiv:2406.03287}, 2024.

\bibitem{yang2022training}
Qu~Yang, Jibin Wu, Malu Zhang, Yansong Chua, Xinchao Wang, and Haizhou Li.
\newblock Training spiking neural networks with local tandem learning.
\newblock {\em Advances in Neural Information Processing Systems}, 35:12662--12676, 2022.

\bibitem{yao2024spike}
Man Yao, Jiakui Hu, Zhaokun Zhou, Li~Yuan, Yonghong Tian, Bo~Xu, and Guoqi Li.
\newblock Spike-driven transformer.
\newblock {\em Advances in neural information processing systems}, 36, 2024.

\bibitem{yao2024spikeChip}
Man Yao, Ole Richter, Guangshe Zhao, Ning Qiao, Yannan Xing, Dingheng Wang, Tianxiang Hu, Wei Fang, Tugba Demirci, Michele De~Marchi, et~al.
\newblock Spike-based dynamic computing with asynchronous sensing-computing neuromorphic chip.
\newblock {\em Nature Communications}, 15(1):4464, 2024.

\bibitem{zenke2021remarkable}
Friedemann Zenke and Tim~P Vogels.
\newblock The remarkable robustness of surrogate gradient learning for instilling complex function in spiking neural networks.
\newblock {\em Neural computation}, 33(4):899--925, 2021.

\bibitem{zhang2022opt}
Susan Zhang, Stephen Roller, Naman Goyal, Mikel Artetxe, Moya Chen, Shuohui Chen, Christopher Dewan, Mona Diab, Xian Li, Xi~Victoria Lin, et~al.
\newblock Opt: Open pre-trained transformer language models.
\newblock {\em arXiv preprint arXiv:2205.01068}, 2022.

\bibitem{zhu2023minigpt}
Deyao Zhu, Jun Chen, Xiaoqian Shen, Xiang Li, and Mohamed Elhoseiny.
\newblock Minigpt-4: Enhancing vision-language understanding with advanced large language models.
\newblock {\em arXiv preprint arXiv:2304.10592}, 2023.

\bibitem{zhu2023spikegpt}
Rui-Jie Zhu, Qihang Zhao, Guoqi Li, and Jason~K Eshraghian.
\newblock Spikegpt: Generative pre-trained language model with spiking neural networks.
\newblock {\em arXiv preprint arXiv:2302.13939}, 2023.

\end{thebibliography}

\medskip

{
\small
}

\newpage
\appendix

\section{The Spike Rate of SNNs}
\label{sec:spike_rate}

This section present the details to get the functional representation of spike rate. First, as indicated in Subsection 3.2 in main text, the kinetic behavior of IF neuron can be represented by Eq.~(\ref{eq:appendix_vt}):
\begin{equation}
\label{eq:appendix_vt}
   \bm{v^{l}(t)} =  v^{l}(t-1)+W^{l}S^{l-1}(t)\theta ^{l-1} -S^{l}(t)\theta ^{l},
\end{equation}
where \(v^{l}(t)\) represent the membrane potential  at time steps \(t\) in the \(i\)-th layers. \(W^{l}\) and \(\theta^l\) are the weight matrix and firing threshold of the IF neuron, respectively. \(S^{l}(t)\) denotes the transmission of discrete spikes at the \(l\)-th layer at time steps \(t\). Note that when \(v^{l}(t-1)+W^{l}S^{l-1}(t)\theta ^{l-1}\) exceeds the threshold \(\theta^{l}\), the IF neuron is fired, and \(S^{l}(t)\) equals to $1$. Otherwise, the IF neuron is muted and \(S^{l}(t)\) equals to $0$.

By accumulating Eq.~(\ref{eq:appendix_vt}) over time steps \(1\) to \(T\), the spike rate $r^{l}(T)$ of layer $l$ can be obtained by Eq.~(\ref{eq3}): 
\begin{equation}
\label{eq3}
    r^{l}(T)=W^{l}r^{l-1}(T)+(-\frac{v^{l}(T)-v^{l}(0)}{T} ).
\end{equation}

It can be seen from the formula that \(r^l\) and \(r^{l-1}\) have a linear relationship, similar to the activation function in ANNs. Therefore, we can map the activation value \(a^l\) of analog neurons in ANNs to \(r^l\) of IF neurons in SNNs. When \(0 < v^l(T) < \theta^l\) and \(W^l r^{l-1}(T) \in (0, \theta^l)\), Eq.~(\ref{eq3}) can be approximated as below in Eq.~(\ref{eq5}):
\begin{equation}
    \label{eq5}
    r^{l}(T)=\frac{\theta ^{l}}{T}\left \lfloor \frac{TW^{l}r^{l-1}(T)+v^{l}(0)}{\theta ^{l}}  \right \rfloor.
\end{equation}

Finally, combining the situation \(W^{l}r^{l-1}(T) \notin (0, \theta ^{l})\), the spike rate \(r^l\) of IF neurons at layer \(l\) can be represented as a continuous step function, as shown in Eq.~(\ref{eq9-1}):
\begin{equation}
\label{eq9-1}
    r^{l}(T)=\text{clip}(\frac{\theta^{l}}{T}\left \lfloor \frac{TW^{l}r^{l-1}(T)+v^{l}(0)}{\theta ^{l}}  \right \rfloor, 0, \theta ^{l} ).
\end{equation}

\section{Algorithm Details of FAS}
\label{sec:alg_details}

The detailed steps of FAS are in Algorithm~\ref{alg:OverallFramework}. Lines~2-4 are the processes of \textbf{\textit{Stage 1}}, addressing QC errors through full-parameter fine-tuning. More specifically, it starts by replacing the activation function with QCFS (Line~2) and fine-tuning the model on the datasets \(D\) and \(\hat{D}\) (Line~3). Subsequently, the weights are transferred from the fine-tuned ANN model to the SNN model (Line~4). Lines~6-11 describe \textbf{\textit{Stage 2}}, which employs a layer-wise and neuron-wise coarse-to-fine calibration optimization strategy. Each layer undergoes layer-wise calibration (Line~7) followed by neuron-wise calibration on mini-batches sampled from \(\hat{D}\) (Line~10).


\begin{algorithm}
\caption{Overall Framework of FAS}
\label{alg:OverallFramework}
\textbf{Input}: Pre-trained ANN model \(f_{ANN}\) , Finetuning Dataset \(D\), Downstream Dataset \(\hat{D}\), Calibration steps \(\rho\).\\
\textbf{Output}:~SNN model~$f_{SNN}$.
\begin{algorithmic}[1]
\STATE /* \textit{\textbf{Stage 1:}} \textbf{Eliminate QC Errors} */
\STATE  Replace Activation function with QCFS in \(f_{ANN}\);   \label{line2}
\STATE Full-parameter fine-tuning with \(D\) and \(\hat{D}\);    \label{line3}
\STATE Copy weights from \(f_{ANN}\) to the SNN \(f_{SNN}\);    \label{line4}
\STATE /* \textit{\textbf{Stage 2:}} \textbf{Eliminate Temporal Error} */   \label{line5}
\FOR{each layer of $f_{SNN}$}   \label{line6}
    \STATE Perform the proposed \textit{layer-wise calibration} strategy for the \(l\)-th layer; \label{line7}
\ENDFOR \label{line8}
\FOR{each minibatch \(\hat{D}'\) sampled from \(\hat{D}\)}  \label{line9}
    \STATE Perform the proposed \textit{neuron-wise calibration} strategy on \(\hat{D}'\);\label{line10}  
\ENDFOR \label{line11}
\STATE \textbf{return} \(f_{SNN}\).
\end{algorithmic}
\end{algorithm}

\section{Detailed Implementation of \textit{NWC}}
\label{sec:detallnwc}
The NWC process is vectorized. In layer $l$, the initial threshold and membrane potential are represented as vectors, i.e., $\theta^l$ and $\nu^l$, respectively. NWC then optimizes the vectors based on the kinetic behavior of IF neurons:

$$
v^l(t) = v^l(t-1) + W^l S^{l-1}(t)\theta^{l-1} - S^l(t)\theta^l.
$$

More specifically, the objective of the NWC process is to minimize the temporal error between the ANN activations $a^l$ and the SNN firing rates $r^l$ using an activation alignment loss:

$$
L^{al} = \text{MSE}\left(a^l, \frac{\sum_{t=1}^\rho S^l \theta^l}{T}\right).
$$

Although parameter optimization is performed at the level of individual neurons, the update process is vectorized. In particular, the thresholds and membrane potentials of all neurons are synchronously updated via vectorized computation. For a given input batch, the gradients for $\theta^l$ and $\nu^l$ are computed in parallel as follows:

$$
\nabla_{\theta^l} L_{all} = \lambda_1 \frac{\partial L^{al}}{\partial \theta^l} + \lambda_2 \frac{\partial L_{logits}}{\partial \theta^l},
$$

$$
\nabla_{\nu^l} L_{all} = \lambda_1 \frac{\partial L^{al}}{\partial \nu^l} + \lambda_2 \frac{\partial L_{logits}}{\partial \nu^l}.
$$
\section{Datasets}
\label{sec:dataset}

This is the supplementary for Section Datasets \& Baselines. For NLU tasks, we chose seven different types of tasks, i.e., six classification and one regression tasks, from the GLUE benchmark. We selected Quora Question Pair (QQP) and Microsoft Research Paraphrase Corpus (MRPC) for classification tasks, and Semantic Textual Similarity Benchmark (STSB) for regression task to evaluate our FAS on similarity and paraphrase tasks. For inference tasks, we opted for MultiGenre Natural Language Inference (MNLI), Question Answering NLI (QNLI), and Recognizing Textual Entailment (RTE) datasets. For single-sentence-based sentiment analysis tasks, we chose Stanford Sentiment Treebank (SST-2). Accuracy is the metric for QQP, MNLI-m, SST-2, QNLI, RTE. MRPC combines accuracy and F1 scores. STS-B uses the Pearson/Spearman correlation.

For NLG task, we chose the following two classic text classification datasets, i.e., Enwik8 and WikiText-103, to evaluate the text generation performance of FAS. Specifically, the Enwik8 dataset is a large-scale text dataset consisting of the first 100 million characters from Wikipedia. It is widely used for character-level language modeling and text generation tasks, providing a challenging benchmark for models due to its extensive and varied content. The Bit-Per-Byte (BPB) metric is commonly employed to assess its performance.
In addition, the WikiText-103 dataset is another comprehensive text dataset derived from Wikipedia articles. It contains over 100 million words and is known for its high-quality, naturally occurring text. WikiText-103 is commonly used for training and evaluating language models, particularly in tasks involving text generation, language modeling, and machine translation. Perplexity (PPL) is the metric of choice for evaluating the performance.

For vision-language tasks, several key benchmarks are widely used. BLINK is a benchmark for multimodal language models, consisting of 14 classic computer vision tasks reformatted into 3,807 multiple-choice questions. It is designed to evaluate visual perception abilities such as relative depth estimation, visual correspondence, forensics detection, and multi-view reasoning. HallusionBench, on the other hand, focuses on image-context reasoning in large visual-language models. It contains 346 images paired with 1,129 expert-crafted questions, assessing logical consistency, response tendencies, and failure modes like language hallucination and visual illusion. MMMU is another crucial benchmark for evaluating multimodal models on advanced, college-level tasks. With 11.5K questions across six core disciplines, 30 subjects, and 183 subfields, it tests perception and reasoning with domain-specific knowledge across 32 heterogeneous image types, including charts, diagrams, maps, and chemical structures.

\section{Baselines}
\label{sec:baseline}
Following Section Datasets \& Baselines, we selected various SOTA baseline models to verify the effectiveness of our FAS on NLU and NLG tasks. 

\paragraph{NLU tasks - } The baselines are as follows:
\begin{itemize}
    \item CBoW~\cite{Wang2018GLUEAM}: CBoW is a simple sentence representation technique that averages the GloVe embeddings of individual words, ignoring syntactic structure and contextual dependencies.
    \item BiLSTM~\cite{Wang2018GLUEAM}: BiLSTM combines LSTM networks with a bidirectional structure to capture both past and future context in sequences
    \item BiLSTM+Attn~\cite{Wang2018GLUEAM}: BiLSTM+Attn combines BiLSTM's sequence understanding with Attention's focus on relevant sentence parts.
    \item GenSen~\cite{subramanian2018learning}: GenSen is a multi-task learning framework that combines diverse objectives to learn general-purpose sentence representations, leading to improved performance on various NLP tasks.
    \item SNN-TextCNN~\cite{Lv2023SpikingCN}: It is a variant of the TextCNN that combines spiking neural networks.
    \item BERT~\cite{Devlin2019BERTPO}: BERT is a bidirectional language model based on the Transformer Encoder-only architecture and an auto-encoding training paradigm.
    \item spikeBERT~\cite{lv2024spikebertlanguagespikformerlearned}: It transfers knowledge from the transformer-based BERT model to the spiking neuron-based architectures with knowledge distillation.
    \item SpikeLM~\cite{xing2024spikelm}: SpikeLM is a novel language model based on SNN that addresses the performance limitations of traditional SNNs in language tasks. By employing an elastic bi-spiking mechanism, SpikeLM achieves competitive performance with deep neural networks on various language tasks while maintaining the energy efficiency of SNNs.
    \item spikingBERT~\cite{Bal2023SpikingBERTDB}: SpikingBERT proposes a novel bioinspired spiking language model. which leverages the average spiking rate of neurons at equilibrium to train a neuromorphic spiking LM using implicit differentiation technique.
    \item OPT~\cite{zhang2022opt}: The OPT (Open Pre-trained Transformers) model is a suite of decoder-only pre-trained transformers ranging from 125 million to 175 billion parameters, designed to match the performance and sizes of GPT-3 models while promoting reproducible and responsible research at scale.
\end{itemize}
\paragraph{NLG tasks - } The selected baselines are as follows:
\begin{itemize}
    \item spikeGPT ~\cite{zhu2023spikegpt}: It explores combining the powerful Transformer architecture with SNN by utilizing linearization and recurrent Transformer blocks..
    \item AstroSNN~\cite{Shen2023AstrocyteEnabledAI}: AstroSNN integrates neuron-astrocyte interactions into the computational paradigm, demonstrating broad applicability across various hardware platforms and narrowing the gap between biological plausibility and neural modeling.
    \item GPT-2 \cite{radford2019language}: It is a Transformer-based deep learning model that leverages self-attention for text dependency parsing and excels in text generation and understanding post-pre-training on extensive data.
\end{itemize}

\textbf{Vision-Language tasks - } The chosen baselines are as follows:
\begin{itemize}
    \item MiniGPT-4-v2-7B \cite{zhu2023minigpt}: MiniGPT-4 is a vision-language model that aligns a frozen visual encoder with a frozen large language model (Vicuna) using a single projection layer, demonstrating advanced multi-modal capabilities with high computational efficiency.
    \item Qwen-VL-8B \cite{bai2023qwen}: Qwen-VL is a large-scale vision-language model series that enhances the Qwen-LM foundation with visual capabilities, enabling advanced multi-modal tasks like image captioning, question answering, and visual grounding, setting new benchmarks in both generalist and dialog-based tasks.
    \item Claude 3 \cite{sonoda2024diagnostic}: Claude 3 is a family of advanced AI models, including Haiku, Sonnet, and Opus, offering progressively higher intelligence, speed, and cost-efficiency, excelling in tasks like expert knowledge, mathematics, content creation, and multilingual conversation.
    \item LLaVA-1.5-7B \cite{liu2024visual}: LLaVA 1.5 is an advanced vision-language model that builds on the previous LLaVA model by improving performance in image captioning, visual question answering, and other vision-language tasks. It integrates large language models with powerful vision encoders, enhancing its ability to process and understand both text and images. This model also benefits from a more robust training framework, leveraging large, diverse datasets and fine-tuning strategies to increase its generalization capabilities across a wide range of multimodal tasks.
    \item PaliGemma-3B \cite{beyer2024paligemma}: PaliGemma is an open Vision-Language Model (VLM) that is based on the SigLIP-So400m vision encoder and the Gemma-2B language model. It is trained to be a versatile and broadly knowledgeable base model that is effective to transfer. It achieves strong performance on a wide variety of open-world tasks.
\end{itemize}

For ANN-SNN conversion methods, the chosen baselines are as follows:
\begin{itemize}
    \item SPR~\cite{hao2023reducing}: It theoretically establishes the mathematical relationship between residual membrane potential and the specific case of unevenness error, and propose an optimization strategy based on residual membrane potential to reduce unevenness error.
    \item QCFS~\cite{bu2023optimal} : It theoretically analyzes ANN-SNN conversion error and derive the estimated activation function of SNNs. Then it proposes the quantization clipfloor-shift activation function to replace the ReLU activation function in source ANNs, which can better approximate the activation function of SNNs.
    \item COS~\cite{Hao2023BridgingTG}: It proposes a method to judge offset spike based on the residual membrane potential and an optimization method to eliminate conversion errors by shifting the initial membrane potential.
\end{itemize}

\begin{table*}[!ht]
\centering
\small
\caption{Training and fine-tuning hyperparameters for BERT and GPT-2. \(^{*}\) denotes the learning rate for the threshold and initial membrane potential. \(^{**}\) denotes the learning rate for all parameters except the threshold and initial membrane potential. \(^{\dagger}\) denotes 1 epoch for the Enwik8 dataset and 3 epochs for the WikiText-103 dataset.}
\resizebox{1.0\textwidth}{!}{%
\begin{tabular}{@{}l llcccc@{}}
\toprule
\textbf{Stage} & \textbf{Model} & \textbf{Dataset}   & \textbf{Learning Rate$^*$} & \textbf{Learning Rate$^{**}$}&\textbf{BatchSize} & \textbf{Epochs/Tokens}  \\ \midrule
\multirow{5}{*}{\textbf{Stage 1}} & \textbf{BERT}  & WikiText-103   & $0.01$  & $5 \times 10^{-5}$ &32& 3 epochs    \\
& \textbf{BERT}  & Downstream Tasks & $0.01$ - $0.08$  & $2 \times 10^{-5}$ to $6 \times 10^{-5}$ & 32 & 3 epochs     \\
& \textbf{GPT-2} & FineWeb-Edu  & $0.001$    & $9 \times 10^{-5}$  &16& $0.3$ billion tokens for 1 epoch \\
& \textbf{GPT-2} & Downstream Tasks  & $0.005$   & $1.2 \times 10^{-4}$  &8& 1 epochs / 3 epcoh $^\dagger $  \\ 
& \textbf{OPT-7B} & FineWeb-Edu &0.005 & $5 \times 10^{-5}$ & 8 & 15600 step \\
\midrule
\multirow{2}{*}{\textbf{Stage 2}} & \textbf{BERT}  & Downstream Tasks & $0.01$ -$0.06$  & - & 16& 3 epochs   \\
& \textbf{GPT-2} & Downstream Tasks & $0.005$& - &2 & 1 epochs / 3 epcoh $^\dagger $   \\ 
& \textbf{OPT-7B} & FineWeb-Edu & 0.001 &-& 8& 13000 step \\
\bottomrule

\end{tabular}
}
\label{tab:param_setup}
\end{table*}

\section{Experiment Settings}
\label{sec:exp_setup}

This section supplements the section of Datasets \& Baselines. To conserve GPU memory, we employed DeepSpeed's ZeRO-2 optimization, utilizing mixed-precision computation on eight Nvidia RTX 3090 GPUs, each with 24GB of memory. Specifically, for the LLaVA-1.5-7B model we used eight NVIDIA RTX H100 GPUs (80 GB each). For stability, gradient clipping was applied with a threshold of 1. The AdamW optimizer was used throughout. BERT was fully fine-tuned on the WikiText-103 dataset, whereas GPT-2 was trained on 0.3 billion tokens from the FineWeb-Edu dataset~\cite{lozhkov2024fineweb-edu}. For the Paligemma and LLaVA models, we used the LLaVA-Instruct-150K dataset.

For downstream tasks, we utilized the respective task's training dataset. In the absence of a standard split, we followed the convention~\cite{Lv2023SpikingCN}, randomly selecting 10\% of the samples as the test set. The hyperparameters were set as follows: \(\alpha^l = 0.6\) and \(\beta^l = 0.1\) for both BERT and GPT-2, \(\lambda_1 = 1\) and \(\lambda_2 = 0.0012\) for GPT-2, and \(\lambda_1 = 1\) with \(\lambda_2\) ranging from 0.2 to 1 in increments of 0.1, selecting the optimal result. 

More detailed settings of the learning rate and epochs for each task are presented in Table~\ref{tab:param_setup}. More specifically, in stage 1, BERT is initially trained on the WikiText-103 dataset used a threshold learning rate of $0.01$ and other parameters set at $5 \times 10^{-5}$, over 3 epochs. Fine-tuning on downstream tasks adjusted the threshold learning rate between $0.01$ and $0.08$, with other parameters ranging from $2 \times 10^{-5}$ to $6 \times 10^{-5}$. In addition, GPT-2 is trained on the FineWeb-Edu dataset and used a threshold learning rate of $0.001$. Other parameters in training GPR-2 are set at $9 \times 10^{-5}$, covering 0.3 billion tokens in 1 epoch. Fine-tuning on WikiText-103 set the threshold learning rate to $0.005$ and other parameters to $1.2 \times 10^{-4}$. In stage 2, the learning rate of BERT is ranged from $0.01$ to $0.06$, depending on the convention of different downstream tasks. Furthermore, the learning rate of GPT-2 is set to $0.005$ on the WikiText-103 dataset.

\section{More Parameter Studies}
\label{sec:appendix_para_study}

\subsection{Parameter Studies on eight timesteps of BERT model}
\label{sec:addExpBERT}
FAS uses four timesteps, whereas other methods typically use eight. This is because FAS can already achieve superior performance with fewer timesteps. Using fewer timesteps also leads to faster SNN inference speed. Additionally, we conducted experiments using the same number of timesteps (eight). As shown in Table~\ref{table:BERT8timestep}, the results demonstrate that FAS still delivers strong performance across various tasks.

\begin{table}[!ht]
\centering
\caption{Performance Comparison of BERT between ANN and SNN}
\label{table:BERT8timestep}
\begin{tabular}{lcccccc}
\midrule
        & T  & QQP &  QNLI & RTE &MRPC &STS-B \\ \midrule
Bert &N/A &90.71 & 90.66 & 65.70 & 84.07/88.85 &88.64/88.48 \\ 
\midrule
FAS (Bert)    & 8  & 90.42 & 90.43   & 65.86      & 86.10/90.28     & 87.88/87.49     \\ \midrule
\end{tabular}
\end{table}

\subsection{Parameter Studies on \texorpdfstring{\(\rho\)}{rho} of BERT model}
\label{sec:param_rho_bert}
We examine the effect of the hyperparameter calibration steps $\rho$ in \textit{\textbf{Stage 2}} of FAS. Table \ref{tableBEffectBert} shows performance of BERT with different values of $\rho$. It is evident that $\rho$ significantly influences performance. When $\rho=2$, BERT consistently achieves better results across all time steps. Notably, at \( T = 8 \), it reaches an accuracy of $90.94$\%, surpassing the $90.88$\% accuracy of ANN, demonstrating the near-lossless conversion capability of FAS.

\begin{table*}[!ht]
\centering
\small
\caption{
Impact of the parameter \(\rho\) in BERT. Baseline refers to the SNN without Stage 2 optimization. 
}
\resizebox{1\textwidth}{!}{
\begin{tabular}{llllllllll}
\hline
\textbf{Method}  & \textbf{Model} & \textbf{ANN} & \textit{T}=1 & \textit{T}=2 & \textit{T}=4 & \textit{T}=6 & \textit{T}=8 & \textit{T}=16& \textit{T}=32\\
\hline
Baseline  &BERT & 90.88 &81.10&81.97&84.97&85.99&86.63&87.39&87.74           \\  
\textbf{FAS }(\(\rho=1\)) &BERT & 90.88& \textbf{88.96}& 89.27& 89.46& 89.93& 89.97&89.95& 89.99 \\        
\textbf{FAS }(\(\rho=2\)) &BERT & 90.88& 88.63& \textbf{89.62}& \textbf{90.61}& \textbf{90.83}& \textbf{90.94}& \textbf{90.74}&90.57 \\        
\textbf{FAS }(\(\rho=4\)) &BERT& 90.88&62.94&88.52&90.13&90.42&90.63&90.66&\textbf{90.77}     \\        
\textbf{FAS }(\(\rho=6\)) &BERT& 90.88&50.54&86.91&90.12&90.39&90.46&90.65&90.66     \\        
\textbf{FAS }(\(\rho=8\)) &BERT& 90.88&50.54&66.83&89.35&89.88 &90.43&90.55& 90.63     \\        \hline
\end{tabular}
}
\label{tableBEffectBert}
\end{table*}

\subsection{Parameter Studies on \texorpdfstring{$\lambda_1:\lambda_2$}{} of Vision-Language Models}
\label{sec:appendix_lambda_vllm}

We studied the impact of hyperparameters $\lambda_1:\lambda_2$ on vision model, i.e., The results are in the Table~\ref{tab:pailgame3BAppendix}, which demonstrates that the setting for original Table 4 results \(\lambda_1:\lambda_2=1:1\), is the best overall across different tasks. More specifically, this balanced setting leads to higher accuracies on HallusionBench compared to using only one weighting (0:1 or 1:0). That is also the case on BLINK and MMMU. 

\begin{table*}[!ht]
\centering
\small
\caption{Impact of the paramete $\lambda_1:\lambda_2$ on PaliGemma-3B Model}
\begin{tabular}{l c c c c c}
\toprule
\multirow{2}*{ $\lambda_1$ : $\lambda_2$} & \multicolumn{3}{c}{HallusionBench} & \multicolumn{1}{c}{BLINK} & \multicolumn{1}{c}{MMMU} \\
~ & Question Pair Acc. & Figure Acc. & Question Acc.  & Test &  Val \\
\midrule
0:1    & 23.41 & 23.20 & 53.20 & 38.92 & 28.00 \\ 
1:0     & 22.41 & 21.38 & 50.269 & 39.13 & 27.66 \\ 
1:1    & 25.27 & 23.41 & 53.52 & 38.97 & 29.67 \\ 
\bottomrule
\end{tabular}
\vskip -0.1in
\label{tab:pailgame3BAppendix}
\end{table*}

\subsection{Parameter Studies on Different \texorpdfstring{$\lambda_1:\lambda_2$}{} Pairs}
\label{sec:appendix_lambda_diff}

For parameter \(\lambda_1\) and \(\lambda_2\), we have results for additional pairs shown in Table~\ref{table:ablationlambda}, which did not yield any improvements.  As the pair 1:0.0012 leads to the best performance, it was used in all subsequent experiments. Extreme settings like, 10:1 yield 17.62, degrade calibration.

\begin{table}[ht]
\centering
\caption{Impact of the paramete $\lambda_1$,~$\lambda_2$ on GPT-2 model}
\label{table:ablationlambda}
\resizebox{1\textwidth}{!}{
\begin{tabular}{lccccccccccc}
\midrule
        &    & \multicolumn{10}{c}{$\lambda_1:\lambda_2$} \\ \midrule
        & T  & 1:10 & 10:01 & 1:2 & 2:1 & 1:1&1:0.0012 &1:0.1&0.1:1&1:0.5&0.5:1\\ \midrule
GPT     & 8  & 17.07 &17.62  &17.04  &17.11   &  17.14  &17.02 & 17.05& 17.04&17.06& 17.12\\ \midrule
\end{tabular}
}
\end{table}

\section{More Efficacy Studies}
\label{sec:appendix_efficacy_study}

\subsection{Efficacy Studies on Approximating GELU function}
\label{sec:appendix_app_gelu}

QCFS can effectively approximate GELU, as shown in the experiments below. We trained a ReLU-activated BERT on the Fineweb-edu dataset, then converted this model to SNNs, along with a vanilla GELU-based BERT, both using FAS. The converted GELU BERT performed comparably to the ReLU counterpart across five GLUE tasks (e.g., QQP: 90.38 vs. 90.35), with performance differences of less than 0.5\%. The comparison shows that FAS can effectively replace GELU without causing a significant performance drop.

\begin{table}[ht]
\centering
\caption{Comparison of Performance of Models with ReLU and GELU Activations}
\begin{tabular}{lccccc}
\midrule
        &     \multicolumn{5}{c}{BERT} \\ \midrule
         & QQP &  QNLI & RTE     &MRPC    &STS-B \\ \midrule
GELU BERT   & 90.71 & 90.66   & 65.70  & 84.07/88.85  & 88.64/88.48 \\ 
Relu BERT      & 90.66 & 90.91   & 66.22  & 86.94/90.61     & 88.62/88.40   \\ \midrule
FAS (GELU BERT)  & 90.38 & 90.13   & 66.06      & 86.02/90.22     & 87.46/87.29     \\ 
FAS (Relu BERT)    & 90.35  & 90.22   & 65.70      & 86.21/90.35     & 88.01/87.87   \\ 
\midrule
\end{tabular}
\end{table}

\subsection{Efficacy Studies on Different Pre-training Qualities}
\label{sec:appendix_pre_train}

We further analyzed LLMs of varying quality by evaluating pre-trained BERT models with 110M and 340M parameters, respectively. The results, demonstrated  in the Table ~\ref{table:Quality_pre}, indicate that higher-quality ANN models lead to higher-quality SNNs. FAS is capable of preserving the performance of these models across different scales.

\begin{table}[!ht]
\centering
\caption{Comparison of BERT Models with Different Pre-training Quality}
\label{table:Quality_pre}
\begin{tabular}{lccccccc}
\hline
         & SST-2 & QNLI & RTE & MRPC  & STS-B    \\
\midrule
Vanilla Bert-110M     & 92.32  & 90.66  & 65.7  & 84.07/88.85 & 88.64/88.48   \\
FAS (Bert-110M) & 91.17  & 90.13  & 66.06 & 86.02/90.22 & 87.46/87.26   \\ \midrule
Vanilla Bert-340M  &  93.00 & 92.29 & 72.92& 89.46/92.65 & 89.55/89.22  \\
FAS (Bert-330M) & 92.06  & 91.55  & 71.66 &89.05/92.22  & 89.69/89.50   \\
\midrule
\end{tabular}
\end{table}

\subsection{Efficacy Studies on the Fine-Tuned ANN-based LLMs and Spiking LLMs}
\label{sec:appendix_fine_tuned_ann}

In this subsection, we compare fine-tuned ANN models after \textbf{Stage 1} with their SNN counterparts. 
Table~\ref{table:fineTunedBERT} shows that, for BERT-based tasks, FAS achieves near-parity with the original fine-tuned models. Also, Table~\ref{table:finetunedOPT} demonstrates that on OPT-7B reasoning benchmarks, FAS remains competitive across multiple time steps (\(T=8,16,32\)), even outperforming the fine-tuned OPT on COPA (84 vs.\ 83 at \(T=8\)) with only minor trade-offs on PIQA (72.74 vs.\ 75.33 at \(T=8\)).
In multimodal evaluations (Table~\ref{table:fineTunedVLLM}), FAS narrows performance gaps, as seen in PaliGemma-3B’s Question Accuracy rising from 52.68 to 53.52.
This systematic validation shows that, when fairly compared to fine-tuned counterparts under identical protocols, FAS effectively preserves model capabilities during conversion.

\begin{table}[ht]
\centering
\caption{Comparison of  Fine-Tune BERT and GPT Models and Their Corresponding SNN Models}
\label{table:fineTunedBERT}
\resizebox{1.0\textwidth}{!}{%
\begin{tabular}{lccccccc|c}
\hline
        & \multicolumn{7}{c|}{BERT} & \multicolumn{1}{c}{GPT} \\
\midrule
        & QQP & MNLI-m & SST-2 & QNLI & RTE & MRPC  & STS-B   & WT \\
\midrule
Vanilla   & 90.71   & 83.91  & 92.32  & 90.66  & 65.7  & 84.07/88.85 & 88.64/88.48  & 16.53 \\
Fine-tuned   & 90.596  & 83.76  & 92.08  & 90.884 & 66.426& 86.27/90.41 & 87.977/87558 & 16.35 \\
FAS     & 90.38   & 82.77  & 91.17  & 90.13  & 66.06 & 86.02/90.22 & 87.46/87.26  & 16.84 \\
\midrule
\end{tabular}
}
\end{table}

\begin{table}[ht]
\centering
\caption{Comparison of Fine-Tuned OPT Models and SNNs}
\label{table:finetunedOPT}
\resizebox{1.0\textwidth}{!}{%
\begin{tabular}{lcccccccc}
\hline
Model      & T    & piqa  & arc\_easy & openbookqa & winogrande & copa  & wsc273 & rte    \\
\midrule
Vanilla (OPT)   & N/A  & 76.26 & 65.57     & 27.6       & 65.43      & 81    & 82.05  & 55.25  \\
Fine-tuned (OPT) & N/A  & 75.33 & 63.64     & 28.80      & 62.98      & 83    & 76.92  & 52.35  \\ \midrule
FAS (OPT) & 8    & 72.74 & 63.97     & 27.60      & 60.30      & 84    & 77.29  & 53.07  \\
FAS (OPT) & 16   & 73.23 & 64.73     & 27.00      & 60.38      & 83    & 77.66  & 55.60  \\
FAS (OPT) & 32   & 74.05 & 64.60     & 27.80      & 60.06      & 82    & 77.29  & 55.23  \\
\hline
\end{tabular}
}
\end{table}

\begin{table}[ht]
\centering
\caption{Comparison of  Fine-Tune PaliGemma and LLaVA and Their Corresponding SNN Models}
\label{table:fineTunedVLLM}
\resizebox{1.0\textwidth}{!}{%
\begin{tabular}{lccccc}
\hline
         & \multicolumn{3}{c}{HallusionBench} & BLINK & MMMU \\
\cline{2-6}
Model    & Question Pair Acc. & Figure Acc. & Question Acc. & Test  & Val   \\
\midrule
Vanilla (PaliGemma) & 22.63  & 21.96   & 51.84   & 38.29 & 32.88 \\
Fine-tuned (PaliGemma) & 25.89  & 24.56   & 52.68   & 39.40 & 32.80 \\
FAS (PaliGemma) & 25.27  & 23.41   & 53.52   & 38.92 & 29.67 \\
\midrule
Vanilla (LLaVA) & 15.31  & 20.87   & 52.78   & 41.22 & 33.66 \\
Fine-tuned (LLaVA)      & 19.57  & 20.04   & 52.7865 & 41.08 & 33.38 \\
FAS (LLaVA) & 18.68  & 19.78   & 51.41   & 40.37 & 31.33 \\
\midrule
\end{tabular}
}
\end{table}

\subsection{Efficacy Studies on the Global Performance of NWC}
\label{sec:appendix_efficacy_on_nwc}
\begin{figure}[htbp]
  \centering
  \begin{minipage}[b]{0.48\linewidth}
    \centering
    \includegraphics[width=\linewidth]{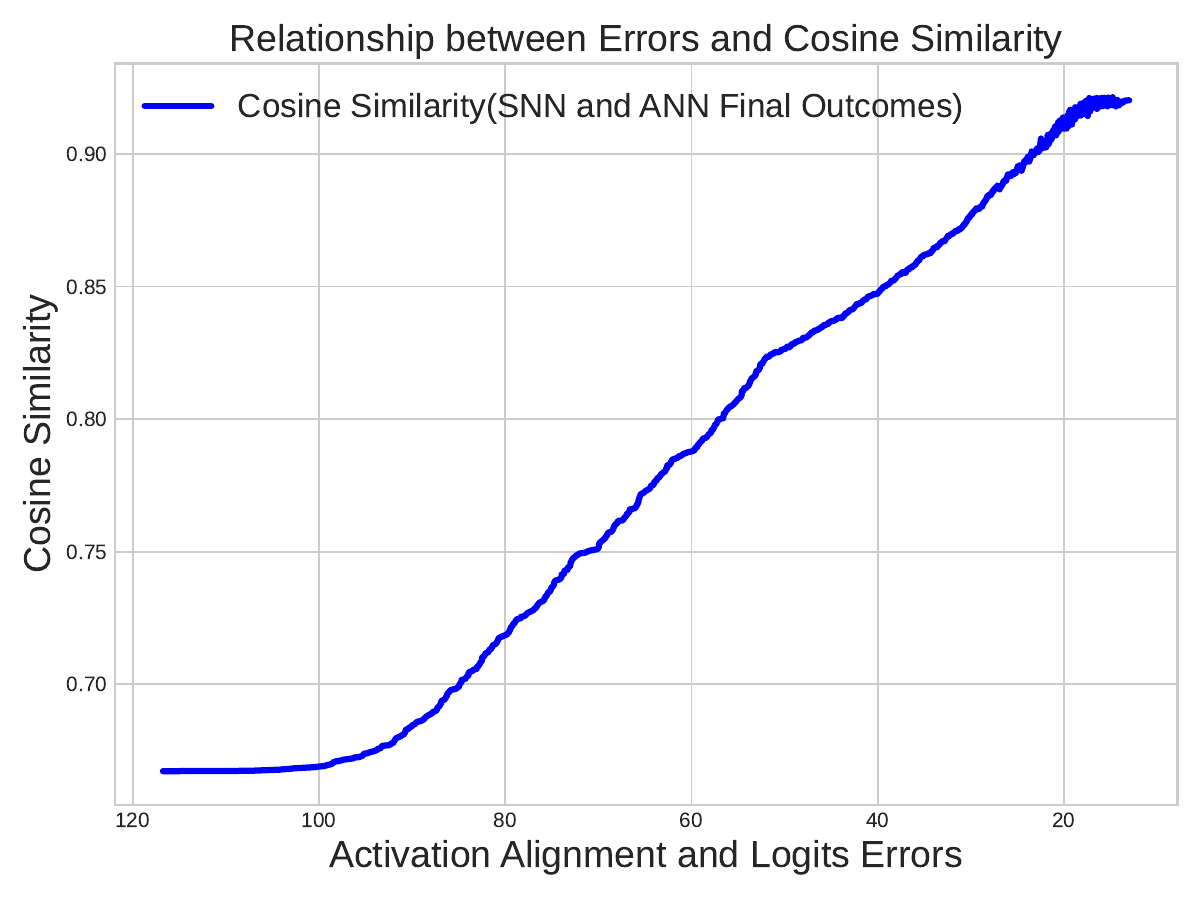}
    \caption{Relationship between Errors and Cosine Similarity}
    \label{fig:error_similar}
  \end{minipage}
  \hfill
  \begin{minipage}[b]{0.48\linewidth}
    \centering
    \includegraphics[width=\linewidth]{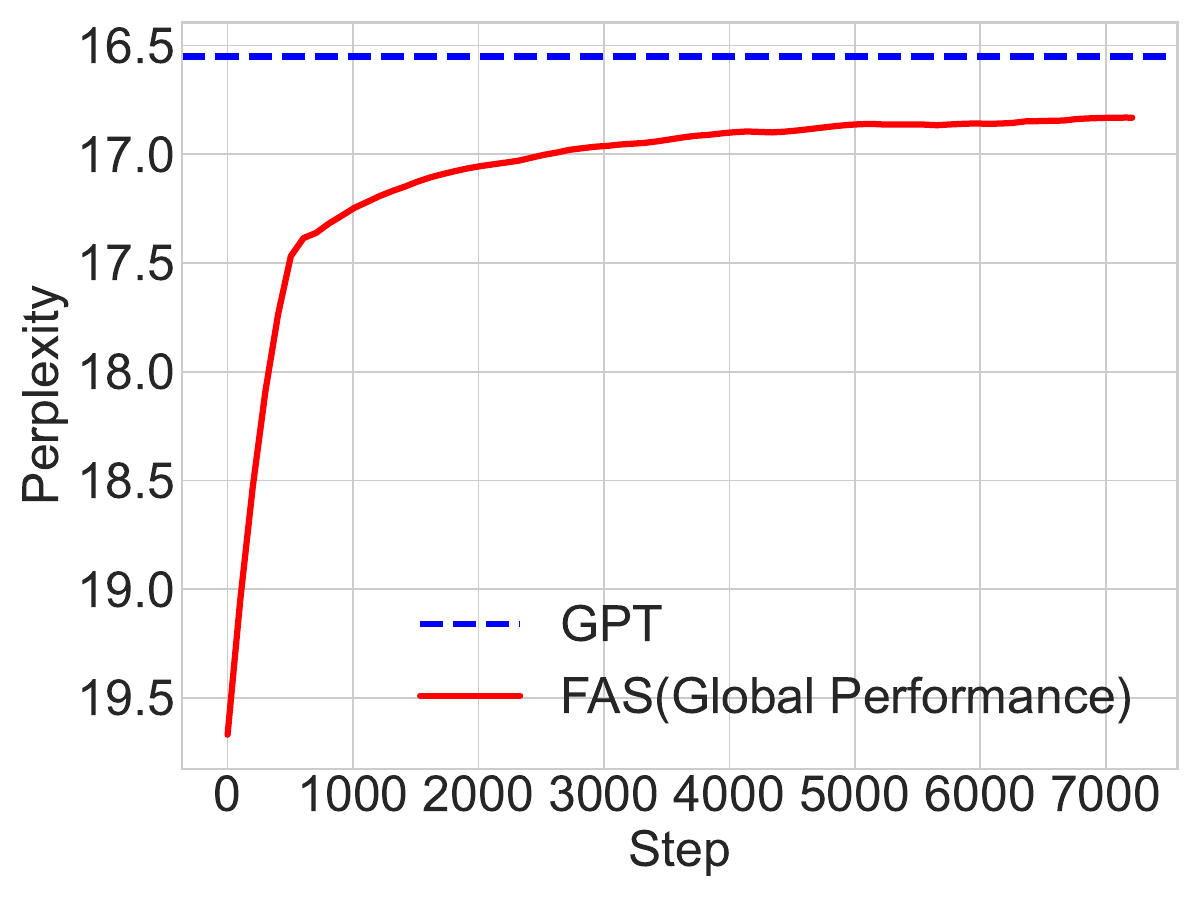}
    \caption{GPT-2 Perplexity Varies with Step}
    \label{fig:perplex_step}
  \end{minipage}
  \label{fig:ab}
\end{figure}

Although NWC operates through localized neuronal-level optimizations, it explicitly aligns SNN activations with their ANN counterparts. It represents the theoretical optimum for minimizing divergence in the final outputs. To validate this, we present cosine similarity measurements between the final outputs of ANN and SNN models in Figure~\ref{fig:error_similar},  which illustrate a progressive convergence of the SNN outputs toward those of the ANN. Additionally, the Figure ~\ref{fig:perplex_step} demonstrates that the perplexity of the SNN model consistently improves during the calibration process, approaching that of the ANN. These results clearly confirm that our calibration strategy successfully translates localized adjustments into global performance gains.

\begin{figure}[!ht]
    \centering
    \includegraphics[width=0.6\linewidth]{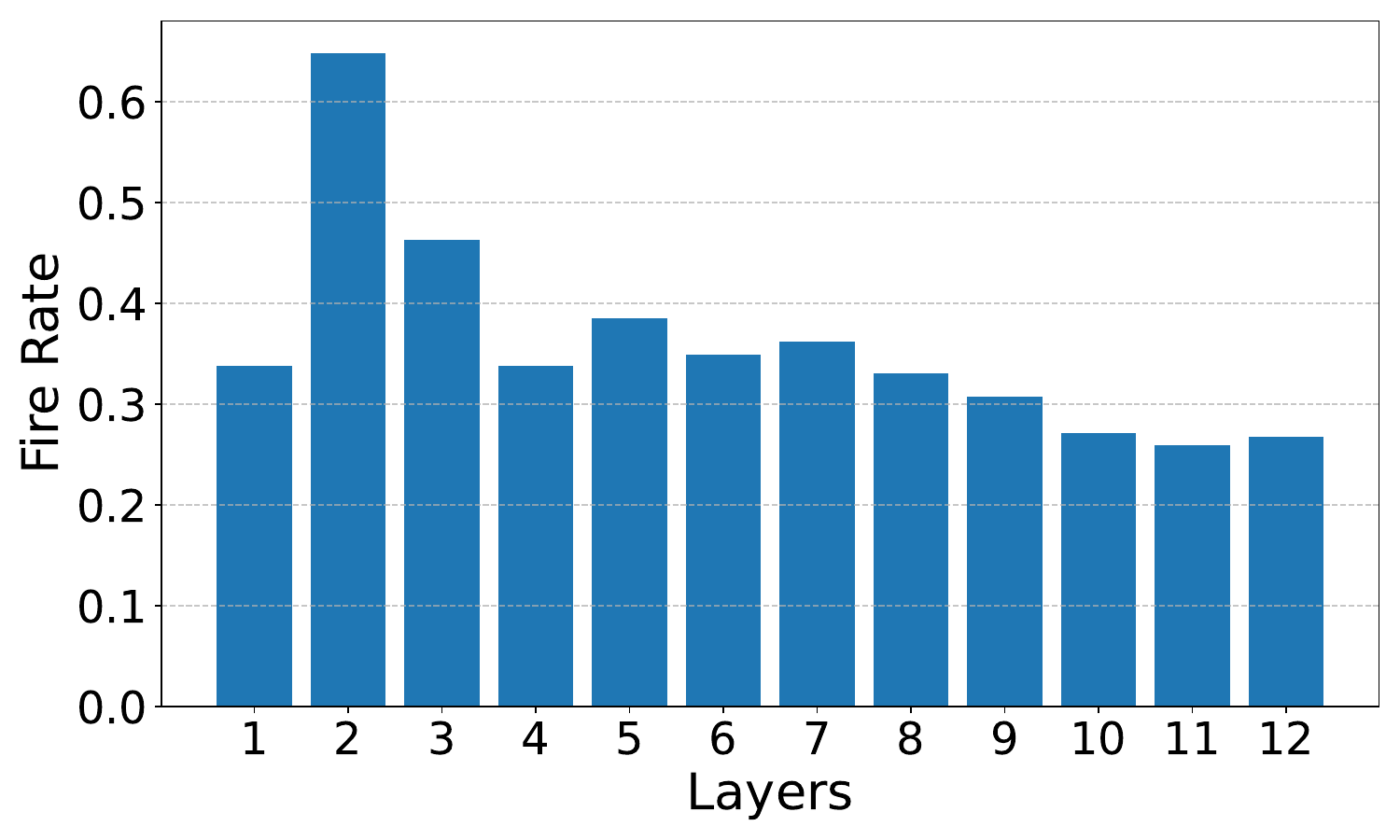}
    \caption{Firing rate visualization of GPT-2.}
    \label{figFireRate}
\end{figure}

\begin{table*}[!ht]
    \centering
    \small
    \caption{Performance and energy usage of converted SNNs relative to ANN for GPT-2. Note that `Per. Deg.' means the `Performance Degradation'.  Also, SRP and COS need an additional 16 time steps to gather the necessary prior information. }
    \resizebox{1.0\textwidth}{!}{%
    \begin{tabular}{l|l|ll|ll|ll|ll}
         \hline
         \textbf{Model} & \textbf{Per. Deg.}& \multicolumn{2}{c|}{\textbf{FAS}} & \multicolumn{2}{c|}{\textbf{QCFS}} & \multicolumn{2}{c|}{\textbf{SRP}} & \multicolumn{2}{c}{\textbf{COS}} \\ 
         \hline \hline

          & & \textbf{Perplexity} & \textbf{Energy (\%)} & \textbf{Perplexity} & \textbf{Energy (\%)} & \textbf{Perplexity} & \textbf{Energy (\%)}& \textbf{Perplexity} & \textbf{Energy (\%)} \\
          \cline{2-10}
        \multirow{10}{*}{\textbf{GPT-2}}& \textbf{ANN (0)} & 16.53 & 100 & 16.53 & 100 & 16.53 & 100 & 16.53 & 100 \\
        \cline{2-10}
        
        & \multirow{2}{*}{\(\bm{\downarrow  (0-1)}\)} & 16.84$_{\textbf{T=16}}$ & 28.19 & &  &  &  &  & \\
        &  & 17.02$_{\textbf{T=8}}$ & 14.21 && & &  &  & \\
      \cline{2-10}
         
        &\(\bm{\downarrow  (1-2)}\) & 17.68$_{\textbf{T=4}}$ & 7.04 &  &  &  &  &  & \\ 
        \cline{2-10}
        
         &\multirow{2}{*}{\(\bm{\downarrow (2-3)}\)} & 19.27$_{\textbf{T=2}}$ & 3.99 &  &  & 19.41$_{\textbf{T=16}}$ & 36.13 &19.16$_{\textbf{T=16}}$ &  35.99 \\
         & &  &  &  &  &  &  &  19.36$_{\textbf{T=8}}$ & 27.11 \\ \cline{2-10}
         
         &\multirow{5}{*}{\(\bm{\downarrow (>3)}\)} & 23.02$_{\textbf{T=1}}$ & 2.19 & 19.64$_{\textbf{T=16}}$  & 17.90 &19.74$_{\textbf{T=8}}$  & 27.08 &19.86$_{\textbf{T=4}}$ & 22.67 \\
        & &  &  & 20.19$_{\textbf{T=8}}$ &8.99  &20.36$_{\textbf{T=4}}$ & 22.53  & 20.81$_{\textbf{T=2}}$ & 20.43 \\
        & &  &  & 21.20$_{\textbf{T=4}}$ & 4.52 &  21.27$_{\textbf{T=2}}$ &  20.22 &  26.38$_{\textbf{T=1}}$ & 19.17 \\
        & &  &  & 22.48$_{\textbf{T=2}}$ & 2.26 & 26.71$_{\textbf{T=1}}$ & 18.93 &  &  \\
        & &  &  & 29.04$_{\textbf{T=1}}$ & 1.00 &  &  &  &  \\

         \hline
    \end{tabular}
}
    \label{energy}
\end{table*}

\section{Analysis of Power \& Energy Efficiency}
\label{sec:energy_analysis}

Since the converted model still involves MAC operations, we focused our comparison solely on the energy consumption of the spiking parts and their corresponding parts in the ANN. Specifically, synaptic operations in SNNs vary depending on spike sparsity with sparse accumulation (AC). In contrast, synaptic operations involving multiplication and accumulation (MAC) in ANNs remain constant within a defined network structure. We measure floating-point AC and MAC operations, using 0.9 pJ per AC and 4.6 pJ per MAC, as reported in~\cite{Li2021AFL}.

To quantitatively assessrgy savings, we compare our converted GPT-2 with their ANN counterparts~\cite{merolla2014million} in terms of performance and energy consumption. Table~\ref{energy} lists the results relative to those of ANN in percentiles for FAS and SOTA ANN-SNN conversation methods QCFS, SRP and COS. On the GPT-2 model, FAS outperforms other ANN-SNN conversion methods in terms of accuracy.  In the case of GPT-2, FAS can achieve similar ANN performance with time steps 2, 4 \& 8. Other methods could not go above the category of ``2-3'' of performance drop, e.g., the increase in perplexity.

Then, we visualize the sparsity of our optimized SNN as shown in Fig.~\ref{figFireRate}, which illustrates the spike rate of all layers of the GPT-2 model using the WikiText-103 dataset with $T = 4$. A spike rate of $1$ means that the numbers of operations in the ANN and SNN are identical.  Fig.~\ref{figFireRate} reveals that the maximum spike rate observed is below 0.64, while the minimum is around 0.25. This suggests that our FAS-generated SNN can significantly reduce the required operations compared to the ANN counterpart.

The analysis of power and energy efficiency demonstrates the following: 
\begin{enumerate}
\item \textbf{FAS achieves SOTA performance across all time steps, surpassing the LLM.}  In particular, the accuracy of FAS with 16 time steps using the BERT model exceeds that of its ANN counterpart and other SOTA methods.
\item \textbf{FAS runs fast, especially under similar energy consumption.} For example, on GPT-2 models, the energy consumption of FAS and QCFC is 7.04\% and 8.99\% under 4 and 8 time steps, respectively. As a result, FAS achieves lower perplexity, and the reduced number of time steps results in faster inference times.
\end{enumerate}

\section{Case Studies on Vision-Language Models}
\label{sec:appendix_case_studies}

\begin{figure}[htbp]
  \centering
  \includegraphics[width=1.0\linewidth]{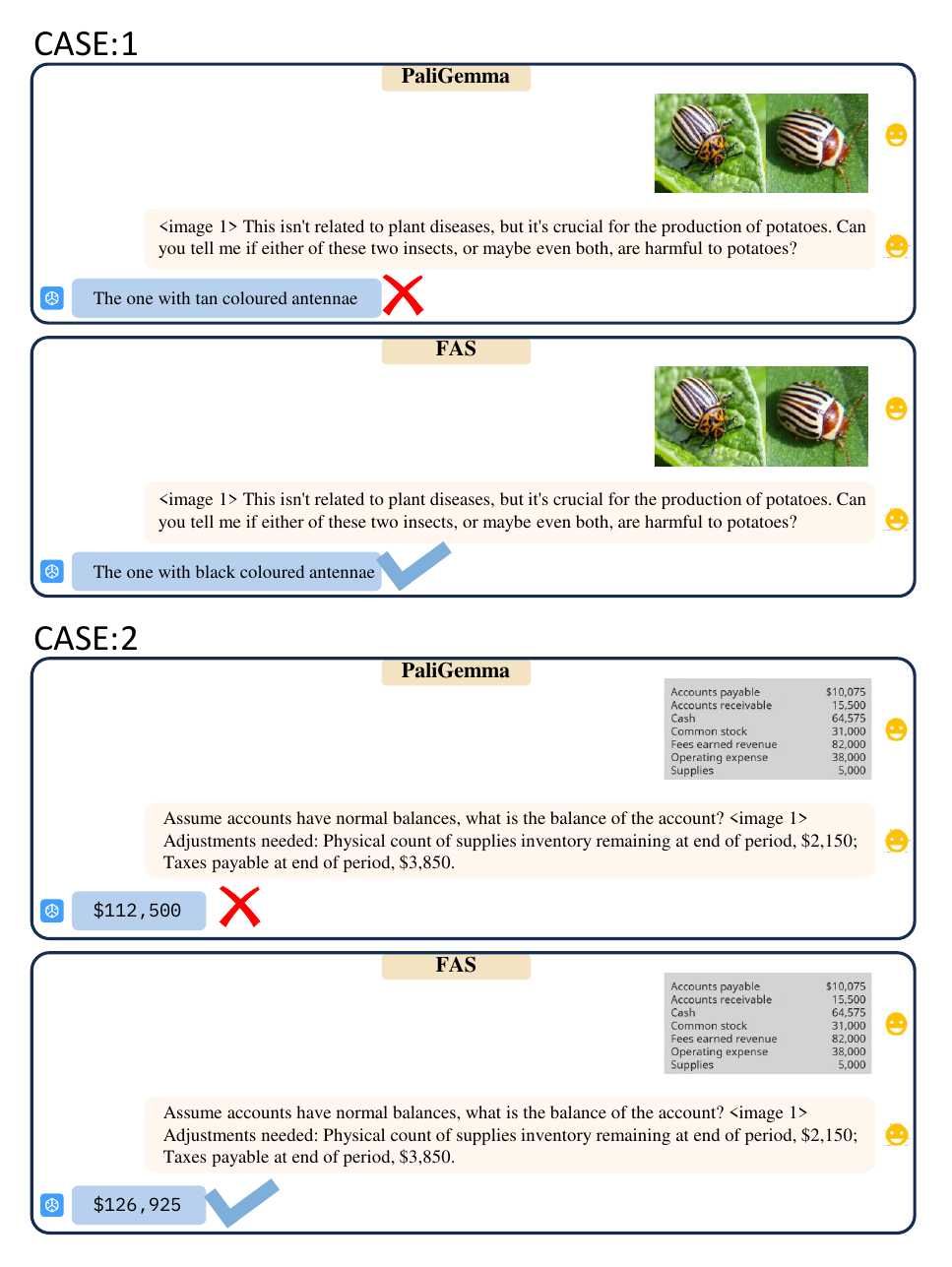}
  \caption{Case Studies on Vision-Language Models}
  \label{fig:exampleVLLM}
\end{figure}

We privide the detailed case studies on the real-world vision-language task. The cases are provided in Figure~\ref{fig:exampleVLLM}. Specifically, in both Case 1 and Case 2, FAS succeeded yet PaliGemma failed. The cases indicate the good performance of FAS and its strong practical viability.

\end{document}